*Article*

# Four-dimensional Gait Surfaces for A Tilt-rotor – Two Color Map Theorem

**Zhe Shen *, Yudong Ma and Takeshi Tsuchiya**

Department of Aeronautics and Astronautics, The University of Tokyo, Tokyo 1138654, Japan;
* Correspondence: zheshen@g.ecc.u-tokyo.ac.jp

**Abstract:** This article presents the four-dimensional surfaces which guide the gait plan for a tilt-rotor. The previous gaits analyzed in the tilt-rotor research are inspired by animals; no theoretical base backs the robustness of these gaits. This research deduces the gaits by diminishing the adverse effect of the attitude of the tilt-rotor for the first time. Four-dimensional gait surfaces are subsequently found, on which the gaits are expected to be robust to the attitude. These surfaces provide the region where the gait is suggested to be planned. However, a discontinuous region may hinder the gait plan process while utilizing the proposed gait surfaces. 'Two Color Map Theorem' is then established to guarantee the continuity of each gait designed. The robustness of the typical gaits on the gait surface, obeying Two Color Map Theorem, is demonstrated by comparing the singular curves in attitude with the gaits not on the gait surface. The result shows that the gaits on the gait surface receive wider regions of the acceptable attitudes.

## 1. Introduction

In stabilizing a quadrotor, feedback linearization is favored for its unique character in linearizing the nonlinear system [1–5]; several degrees of freedom can be subsequently controlled independently by this method. These degrees of freedom can be selected as attitude and altitude [4,6,7], position and yaw angle [8–11], attitude only [12], etc. The reason for not assigning all six degrees of freedom as independent controlled variables is that the number of inputs in a conventional quadrotor is four, marking the maximum number of degrees of freedom to be independently controlled [13] less than the entire degrees of freedom.

Typical feedback linearization requires an invertible decoupling matrix [14,15], which is always satisfied for a quadrotor with attitude-altitude independent output choice. While the position-yaw independent output choice witnesses a singular decoupling matrix for some attitudes [11,16].

On the other hand, some research stabilize Ryll's tilt-rotor [17–19], a novel UAV with eight inputs, by controlling all degrees of freedom independently and simultaneously. The relevant decoupling matrices with six inputs and eight inputs have been proved invertible, marking the feasibility of feedback linearization.

These controllers, however, can command the tilting-angles of the tilt-rotor to change over-intensively [17,20,21], not expected in application.

With this concern, our previous research sets the magnitudes of the thrusts the only four inputs, assigned by the subsequent united control rule [22]. While the four tilting-angles are defined in gait plan beforehand, which is totally not influenced by the control rule.

Undeniably, the maximum number of degrees of freedom that can be independently controlled decreases to four, less than the number of the entire degrees of freedom (six);

attitude and altitude are selected as the only degrees of freedom to be directly stabilized. The remaining degrees of freedom, X and Y in position, are tracked by adjusting the attitude properly based on a modified attitude-position decoupler [23]. While the conventional attitude-position decoupler [7,24–26] for a quadrotor no longer works for a tilt-rotor.

Note that the relevant decoupling matrix can be singular for some attitudes while adopting some gaits [22]. Various animal-inspired gaits are subsequently analyzed and modified to avert a non-invertible decoupling matrix by scaling [27–29], which has been proved to be a valid approach to modify a gait. The modified gaits show wider regions of the acceptable attitudes in the roll-pitch diagram, indicating the enhanced robustness.

However, no research guides the gait plan, considering the robustness, so far, except adopting an exist gait (e.g., animal-inspired gait [23,27]). Indeed, deducing the explicit relationship between the attitude and the singularity of the decoupling matrix can be cumbersome since they are tangled in a highly non-linear way.

This article articulates the influence of the attitude to the singularity of the decoupling matrix by little attitude approximation. The deduced gaits are robust to the change of the attitude; restricted disturbances in attitude will not change the singularity of the decoupling matrix while adopting the deduced gaits. The resulting acceptable gaits lie on the four-dimensional gait surface.

To avoid the discontinuous change in the tilting-angles, the gait is required to move continuously along all the four dimensions of the four-dimensional gait surface. Two Color Map Theorem is subsequently developed to further guide the gait plan to design a continuous gait, which is inspired by the well-known Four Color Map Theorem [30,31].

The acceptable attitudes of four typical gaits on the different parts of the four-dimensional gait surface are analyzed in the roll-pitch diagram. The results are compared with the relevant biased gaits, locating outside the four-dimensional gait surfaces. It shows that the robustness of the gait is improved if the gait is planned on the nearby four-dimensional gait surface.

The rest of this article is organized as follows. Section 2 reviews the necessary condition to receive the invertible decoupling matrix for our tilt-rotor. Section 3 thoroughly deduces the acceptable gaits considering the robustness. The resulting four-dimensional gait surfaces are also visualized in the same section. Section 4 proposes Two Color Map Theorem to guide a continuous gait plan. The verification of the robustness of the proposed gaits is analyzed in the roll-pitch diagram in Section 5. Finally, Section 6 addresses the conclusions and discussions.

## 2. Invertible Decoupling Matrix: Necessary Condition

The target tilt-rotor analyzed in this research is sketched in Figure 1 [22]. This structure was initially put forward by Ryll [13].

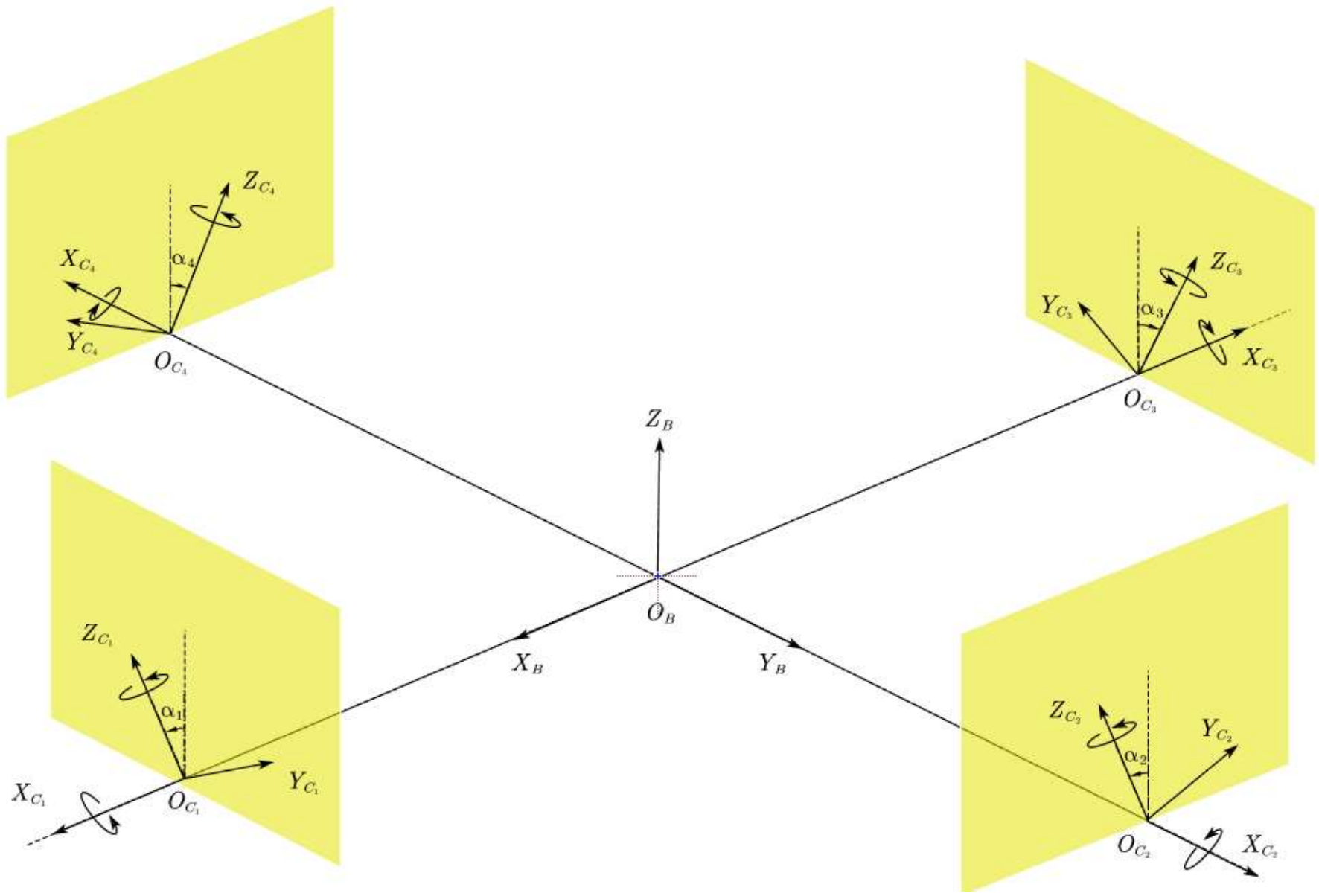

**Figure 1.** The sketch of Ryll's tilt-rotor.

The frames introduced in the dynamics of this tilt-rotor are earth frame $\mathcal{F}_E$, body-fixed frame $\mathcal{F}_B$, and four rotor frames $\mathcal{F}_{C_i}(i$=1,2,3,4), each of which is fixed on the tilt motor mounted at the end of each arm. Rotor 1 and 3 are assumed to rotate clockwise along $Z_{C_1}$ and $Z_{C_3}$. While rotor 2 and 4 are assumed to rotate counter-clockwise along $Z_{C_2}$ and $Z_{C_4}$.

As can be seen in Figure 1, the thrusts can be assigned to vector along the directions one the highlighted yellow planes (tilting). This is completed by changing the tilting-angles marked by $\alpha_1, \alpha_2, \alpha_3, \alpha_4$ in Figure 1.

Ryll proved that the decoupling matrix in feedback linearization, programmed to independently stabilize each degree of freedom, is invertible [13]. It guarantees that the dynamic inversion can be conducted throughout the entire flight for any attitude. This approach, however, introduces over-intensive changes in the tilting-angles.

To overcome this obstacle, an alternative selection of the independent controlled degrees of freedom, attitude and altitude, is given [22] while utilizing the magnitudes as the only inputs. Define the combination of the tilting-angle $(\alpha_1, \alpha_2, \alpha_3, \alpha_4)$ as gait. The tilting-angles $(\alpha_1, \alpha_2, \alpha_3, \alpha_4)$ are planed beforehand as a separate process (gait plan), not influenced by the subsequent feedback linearization. The only inputs resulted from a united control rule are the magnitudes of the four thrusts.

Interestingly, this selection of independent controlled variables introduces no singular decoupling matrix in a conventional quadrotor, which can be regarded as a special case for a tilt-rotor, e.g., $\alpha_1 = 0, \alpha_2 = 0, \alpha_3 = 0, \alpha_4 = 0$. However, it introduces the singular decoupling matrix in a tilt-rotor for some tilting angles given some attitudes.

Specifically, the necessary condition [22] to receive an invertible decoupling matrix in our tilt-rotor is given by

$$
\begin{aligned}
&1.000 \cdot c1 \cdot c2 \cdot c3 \cdot s4 \cdot s\theta - 1.000 \cdot c1 \cdot s2 \cdot c3 \cdot c4 \cdot s\theta - 2.880 \cdot c1 \\
&\cdot c2 \cdot s3 \cdot s4 \cdot s\theta + 2.880 \cdot c1 \cdot s2 \cdot s3 \cdot c4 \cdot s\theta - 2.880 \cdot s1 \cdot c2 \cdot c3 \\
&\cdot s4 \cdot s\theta + 2.880 \cdot s1 \cdot s2 \cdot c3 \cdot c4 \cdot s\theta - 1.000 \cdot s1 \cdot c2 \cdot s3 \cdot s4 \cdot s\theta \\
&+ 1.000 \cdot s1 \cdot s2 \cdot s3 \cdot c4 \cdot s\theta + 4.000 \cdot c1 \cdot c2 \cdot c3 \cdot c4 \cdot c\phi \cdot c\theta \\
&+ 5.592 \cdot c1 \cdot c2 \cdot c3 \cdot s4 \cdot c\phi \cdot c\theta - 5.592 \cdot c1 \cdot c2 \cdot s3 \cdot c4 \cdot c\phi \cdot c\theta \\
&+ 5.592 \cdot c1 \cdot s2 \cdot c3 \cdot c4 \cdot c\phi \cdot c\theta - 5.592 \cdot s1 \cdot c2 \cdot c3 \cdot c4 \cdot c\phi \cdot c\theta \\
&+ 1.000 \cdot c1 \cdot c2 \cdot s3 \cdot c4 \cdot s\phi \cdot c\theta + 0.9716 \cdot c1 \cdot c2 \cdot s3 \cdot s4 \cdot c\phi \cdot c\theta \\
&- 2.000 \cdot c1 \cdot s2 \cdot c3 \cdot s4 \cdot c\phi \cdot c\theta + 0.9716 \cdot c1 \cdot s2 \cdot s3 \cdot c4 \cdot c\phi \cdot c\theta \\
&- 1.000 \cdot s1 \cdot c2 \cdot c3 \cdot c4 \cdot s\phi \cdot c\theta + 0.9716 \cdot s1 \cdot c2 \cdot c3 \cdot s4 \cdot c\phi \cdot c\theta \\
&- 2.000 \cdot s1 \cdot c2 \cdot s3 \cdot c4 \cdot c\phi \cdot c\theta + 0.9716 \cdot s1 \cdot s2 \cdot c3 \cdot c4 \cdot c\phi \cdot c\theta \\
&+ 2.880 \cdot c1 \cdot c2 \cdot s3 \cdot s4 \cdot s\phi \cdot c\theta + 2.880 \cdot c1 \cdot s2 \cdot s3 \cdot c4 \cdot s\phi \cdot c\theta \\
&- 0.1687 \cdot c1 \cdot s2 \cdot s3 \cdot s4 \cdot c\phi \cdot c\theta - 2.880 \cdot s1 \cdot c2 \cdot c3 \cdot s4 \cdot s\phi \cdot c\theta \\
&+ 0.1687 \cdot s1 \cdot c2 \cdot s3 \cdot s4 \cdot c\phi \cdot c\theta - 2.880 \cdot s1 \cdot s2 \cdot c3 \cdot c4 \cdot s\phi \cdot c\theta \\
&- 0.1687 \cdot s1 \cdot s2 \cdot c3 \cdot s4 \cdot c\phi \cdot c\theta + 0.1687 \cdot s1 \cdot s2 \cdot s3 \cdot c4 \cdot c\phi \\
&\cdot c\theta - 1.000 \cdot c1 \cdot s2 \cdot s3 \cdot s4 \cdot s\phi \cdot c\theta + 1.000 \cdot s1 \cdot s2 \cdot c3 \cdot s4 \cdot s\phi \\
&\cdot c\theta \\
&\neq 0
\end{aligned}
\tag{1}
$$

where $s\Lambda = \sin(\Lambda)$ and $c\Lambda = \cos(\Lambda)$, $si = \sin(\alpha_i)$, $ci = \cos(\alpha_i)$, $(i = 1,2,3,4)$, $\phi$ and $\theta$ are roll angle and pitch angle, respectively.

Since the attitude is not predictable in the gait plan process, the left side of Formula (1) cannot be achieved in gait plan. Our precious research [23] makes zero attitude approximation to Formula (1): substituting $\phi = 0$ and $\theta = 0$ into Formula (1) yields

$$
\begin{aligned}
&4.000 \cdot c1 \cdot c2 \cdot c3 \cdot c4 + 5.592 \\
&\cdot (+c1 \cdot c2 \cdot c3 \cdot s4 - c1 \cdot c2 \cdot s3 \cdot c4 + c1 \cdot s2 \cdot c3 \cdot c4 - s1 \cdot c2 \cdot c3 \cdot c4) \\
&+ 0.9716 \\
&\cdot (+c1 \cdot c2 \cdot s3 \cdot s4 + c1 \cdot s2 \cdot s3 \cdot c4 + s1 \cdot c2 \cdot c3 \cdot s4 + s1 \cdot s2 \cdot c3 \cdot c4) \\
&+ 2.000 \cdot (-c1 \cdot s2 \cdot c3 \cdot s4 - s1 \cdot c2 \cdot s3 \cdot c4) + 0.1687 \\
&\cdot (-c1 \cdot s2 \cdot s3 \cdot s4 + s1 \cdot c2 \cdot s3 \cdot s4 - s1 \cdot s2 \cdot c3 \cdot s4 + s1 \cdot s2 \cdot s3 \cdot c4) \\
&\neq 0.
\end{aligned}
\tag{2}
$$

Formula (2) contains the tilting-angles only, making the verification in the gait plan process possible. Several animal-inspired gaits were then evaluated by Formula (2), and received satisfying tracking result [27].

This result (Formula (2)) from zero attitude approximation, however, discards the influence of the attitude. The effect of the disturbance of roll angle and pitch angle, which may violate Formula (1) cannot be traced from Formula (2).

From this point, though Formula (2) may be useful to determine whether a gait is feasible, it is not suitable to design a robust gait, e.g., design a gait robust to the attitude change. A novel method of planning robust gaits is specified in the next section.

## 3. Four-dimensional Gait Surface

Instead of utilizing zero attitude approximation, make the following near zero attituded approximation: $s\Lambda = \Lambda$, $c\Lambda = 1$. Substituting this near zero attitude approximation into Formula (1) yields

$$R_{\phi}(\alpha_1,\alpha_2,\alpha_3,\alpha_4) \cdot \phi + R_{\theta}(\alpha_1,\alpha_2,\alpha_3,\alpha_4) \cdot \theta + R(\alpha_1,\alpha_2,\alpha_3,\alpha_4) \neq 0 \tag{3}$$

where

$$R_{\phi}(\alpha_1,\alpha_2,\alpha_3,\alpha_4) \tag{4}$$

$$= 1.000 \cdot c1 \cdot c2 \cdot s3 \cdot c4 - 1.000 \cdot s1 \cdot c2 \cdot c3 \cdot c4 - 1.000 \cdot c1 \cdot s2 \cdot s3 \cdot s4 + 1.000 \cdot s1 \cdot s2 \cdot c3 \cdot s4 + 2.880 \cdot c1 \cdot c2 \cdot s3 \cdot s4 + 2.880 \cdot c1 \cdot s2 \cdot s3 \cdot c4 - 2.880 \cdot s1 \cdot c2 \cdot c3 \cdot s4 - 2.880 \cdot s1 \cdot s2 \cdot c3 \cdot c4,$$

$$R_\theta(\alpha_1,\alpha_2,\alpha_3,\alpha_4) = 1.000 \cdot c1 \cdot c2 \cdot c3 \cdot s4 - 1.000 \cdot c1 \cdot s2 \cdot c3 \cdot c4 - 1.000 \cdot s1 \cdot c2 \cdot s3 \cdot s4 + 1.000 \cdot s1 \cdot s2 \cdot s3 \cdot c4 - 2.880 \cdot c1 \cdot c2 \cdot s3 \cdot s4 + 2.880 \cdot c1 \cdot s2 \cdot s3 \cdot c4 - 2.880 \cdot s1 \cdot c2 \cdot c3 \cdot s4 + 2.880 \cdot s1 \cdot s2 \cdot c3 \cdot c4, \quad (5)$$

$$R(\alpha_1,\alpha_2,\alpha_3,\alpha_4) = 4.000 \cdot c1 \cdot c2 \cdot c3 \cdot c4 + 5.592 \cdot c1 \cdot c2 \cdot c3 \cdot s4 - 5.592 \cdot c1 \cdot c2 \cdot s3 \cdot c4 + 5.592 \cdot c1 \cdot s2 \cdot c3 \cdot c4 - 5.592 \cdot s1 \cdot c2 \cdot c3 \cdot c4 + 0.9716 \cdot c1 \cdot c2 \cdot s3 \cdot s4 + 0.9716 \cdot c1 \cdot s2 \cdot s3 \cdot c4 + 0.9716 \cdot s1 \cdot c2 \cdot c3 \cdot s4 + 0.9716 \cdot s1 \cdot s2 \cdot c3 \cdot c4 - 2.000 \cdot c1 \cdot s2 \cdot c3 \cdot s4 - 2.000 \cdot s1 \cdot c2 \cdot s3 \cdot c4 - 0.1687 \cdot c1 \cdot s2 \cdot s3 \cdot s4 + 0.1687 \cdot s1 \cdot c2 \cdot s3 \cdot s4 - 0.1687 \cdot s1 \cdot s2 \cdot c3 \cdot s4 + 0.1687 \cdot s1 \cdot s2 \cdot s3 \cdot c4. \quad (6)$$

To diminish the disturbance of the attitude, roll angle and pitch angle on the left side of Formula (3), both $R_\phi(\alpha_1,\alpha_2,\alpha_3,\alpha_4)$ and $R_\theta(\alpha_1,\alpha_2,\alpha_3,\alpha_4)$ are set zero.

Further, to satisfy Formula (3), $R(\alpha_1,\alpha_2,\alpha_3,\alpha_4)$ is required to be non-zero.

In conclusion, the necessary condition to receive an invertible gait considering robustness is

$$R_\phi(\alpha_1,\alpha_2,\alpha_3,\alpha_4) = 0, \quad (7)$$

$$R_\theta(\alpha_1,\alpha_2,\alpha_3,\alpha_4) = 0, \quad (8)$$

$$R(\alpha_1,\alpha_2,\alpha_3,\alpha_4) \neq 0. \quad (9)$$

Since $R_\phi(\alpha_1,\alpha_2,\alpha_3,\alpha_4)$, $R_\theta(\alpha_1,\alpha_2,\alpha_3,\alpha_4)$, and $R(\alpha_1,\alpha_2,\alpha_3,\alpha_4)$ are highly non-linear, the numerical results will be calculated only.

Evenly dividing the range $\alpha_1 \in [-\pi/2,\pi/2]$ into 16 pieces generates 17 grids, $-\pi/2 + (i-1) \cdot \pi/16, (i=\overline{1,17})$. Only these $\alpha_1$ on the grids are to be calculated. Similarly, only $\alpha_2 = -\pi/2 + (i-1) \cdot \pi/16, (i=\overline{1,17})$ are to be calculated. With this configuration, there are altogether $17 \times 17$ $(\alpha_1,\alpha_2)$ on the grids.

For each $(\alpha_1,\alpha_2)$, the corresponding $(\alpha_3,\alpha_4)$ within the range $\alpha_3 \in [-\pi/2,\pi/2]$, $\alpha_4 \in [-\pi/2,\pi/2]$ are found based on Equation (7) and (8) with the aid of Mathematica. The solver used is NSolve (Precision Set=10).

Figure 2 plots the number of the roots of $(\alpha_3,\alpha_4)$ for each given $(\alpha_1,\alpha_2)$ on the grids.

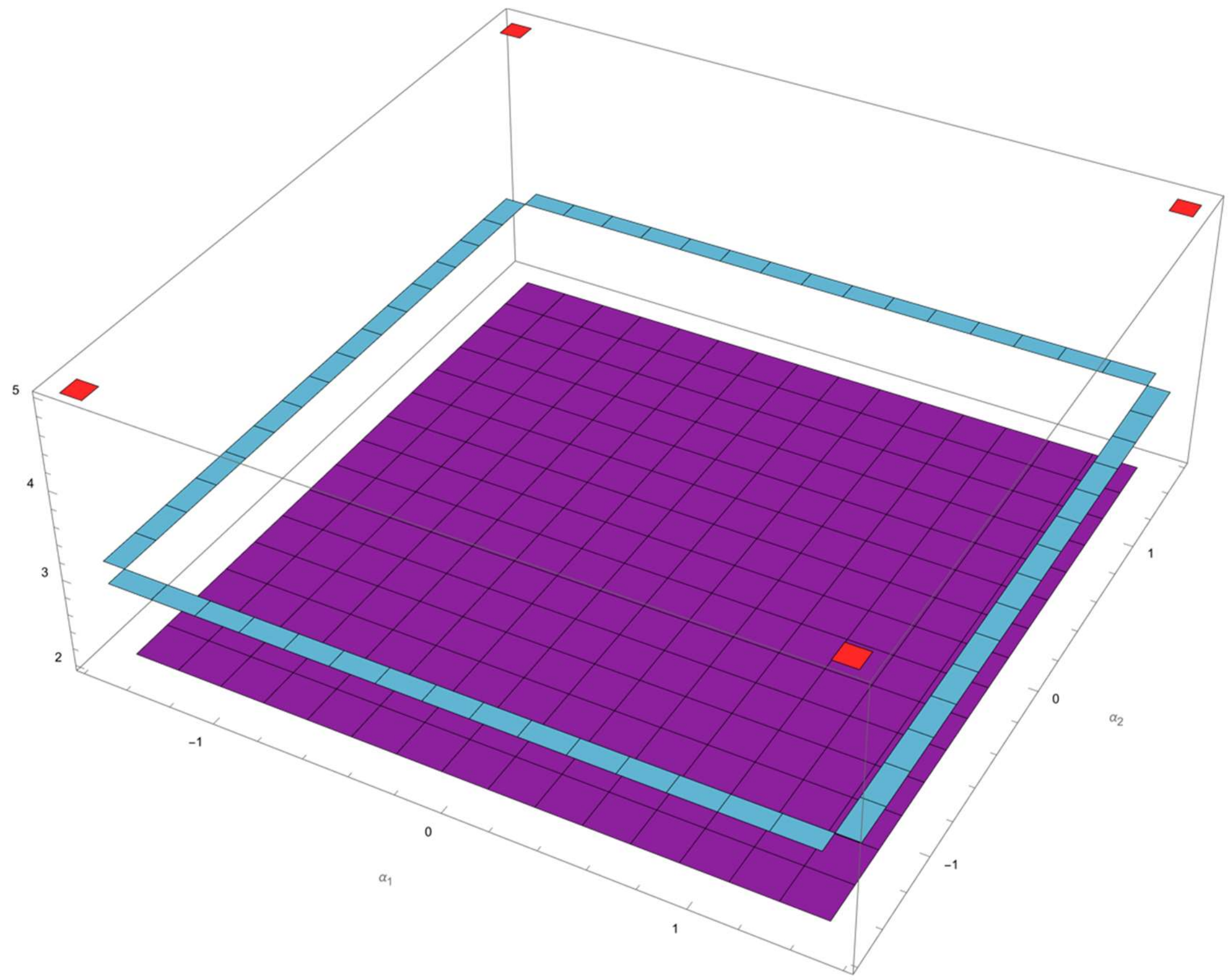

**Figure 2.** Distribution of the number of roots of ($\alpha_3$,$\alpha_4$) for different ($\alpha_1$,$\alpha_2$).

It can be seen that the number of roots of ($\alpha_3$,$\alpha_4$) is 2 in the middle, $\alpha_1 \in (-\pi/2,\pi/2) \cap \alpha_2 \in (-\pi/2,\pi/2)$. Each corner, $\alpha_1 \in \{-\pi/2,\pi/2\} \cap \alpha_2 \in \{-\pi/2,\pi/2\}$, returns 5 roots to the corresponding ($\alpha_3$,$\alpha_4$). Each of the rest ($\alpha_1$,$\alpha_2$) on the side returns 3 roots of ($\alpha_3$,$\alpha_4$).

Discarding ($\alpha_1$,$\alpha_2$) on the side and the corner, we only discuss the ($\alpha_1$,$\alpha_2$) in the middle, where there are two roots for each given ($\alpha_1$,$\alpha_2$).

Figure 3 and Figure 4 plot $\alpha_3$ and $\alpha_4$, respectively, in one of two roots of ($\alpha_3$,$\alpha_4$).

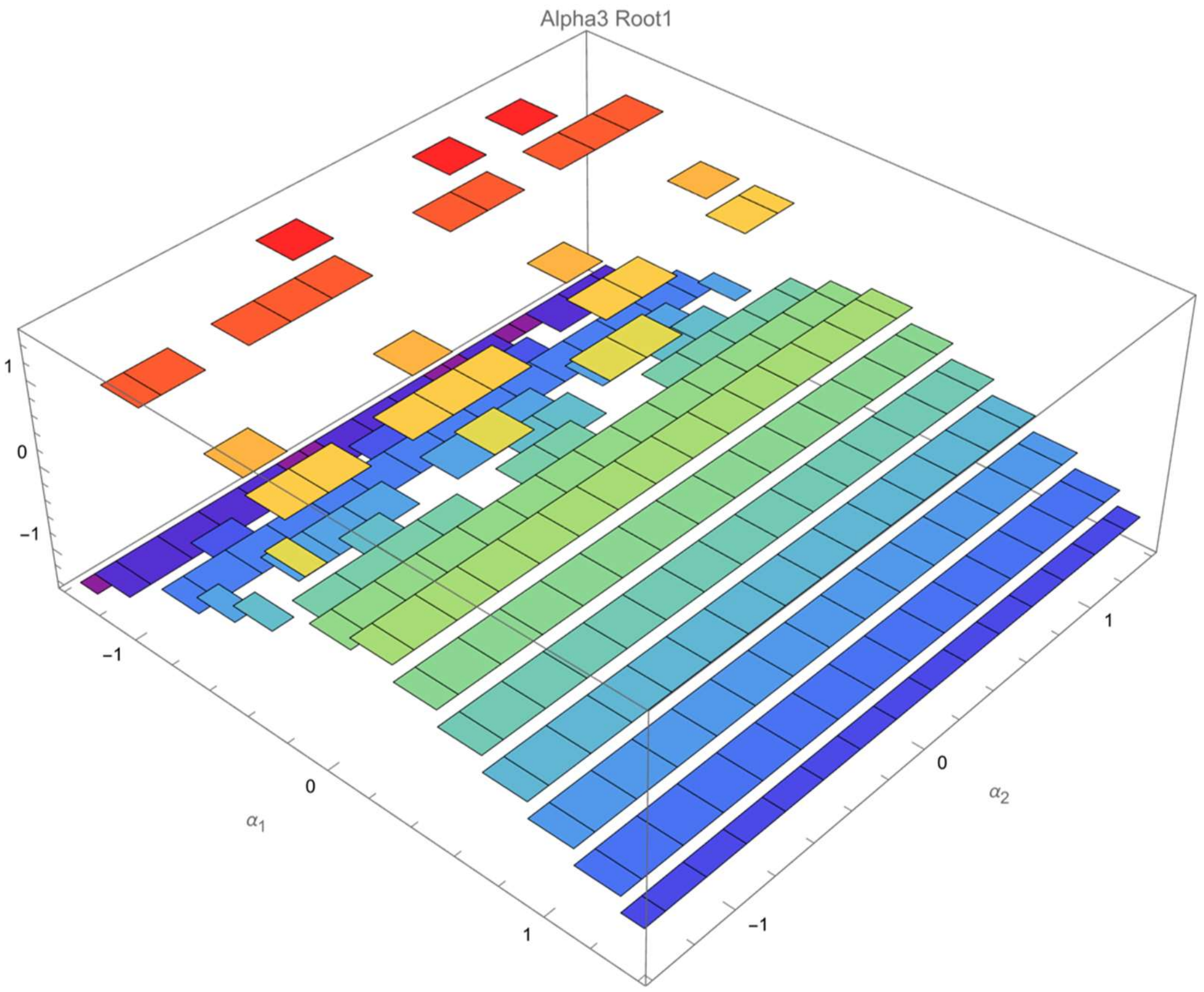

**Figure 3.** The value of $\alpha_3$ in one of the roots of ($\alpha_3$,$\alpha_4$).

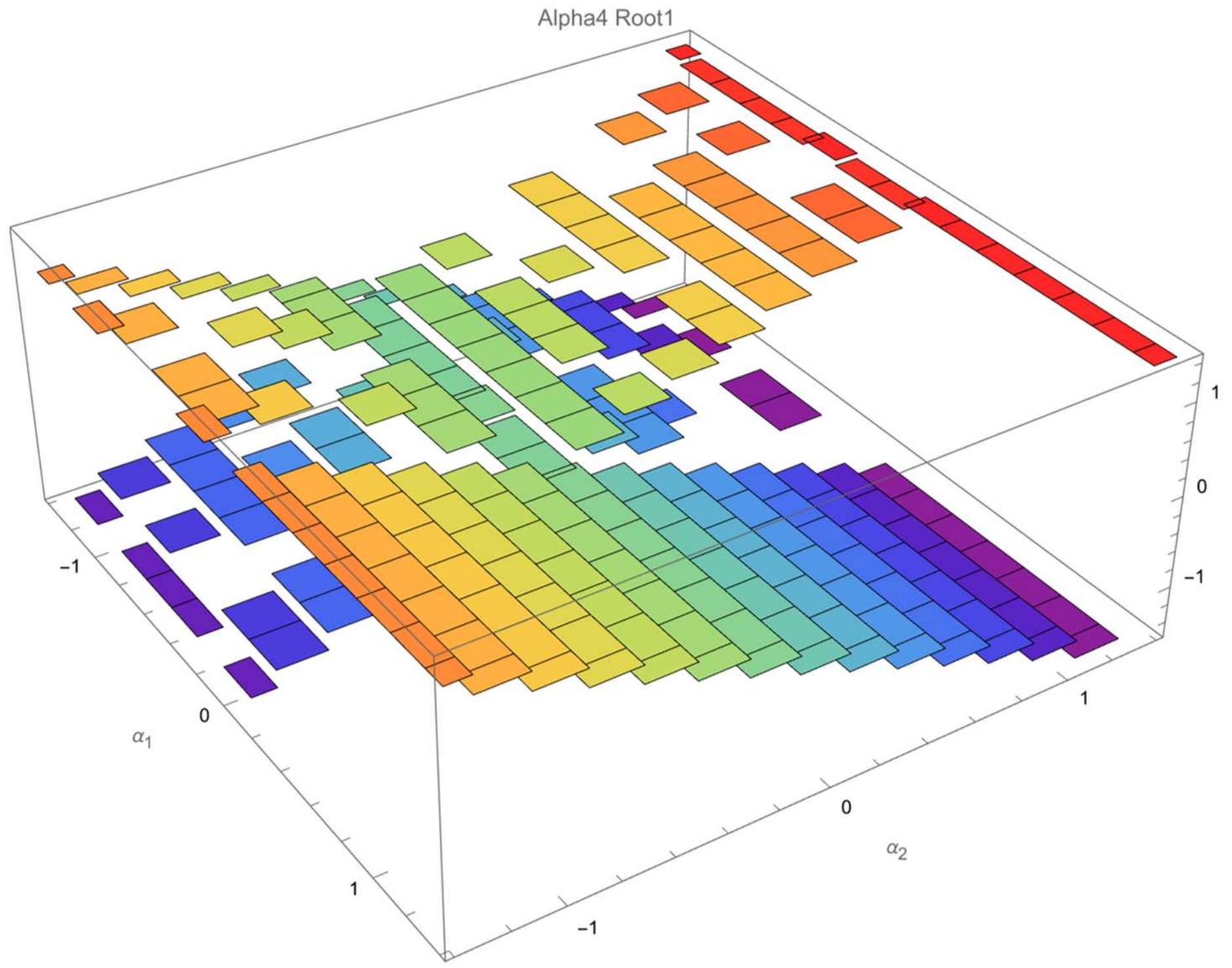

**Figure 4.** The value of $\alpha_4$ in one of the roots of ($\alpha_3$,$\alpha_4$).

Figure 5 and Figure 6 plot $\alpha_3$ and $\alpha_4$, respectively, in the other root of ($\alpha_3$,$\alpha_4$).

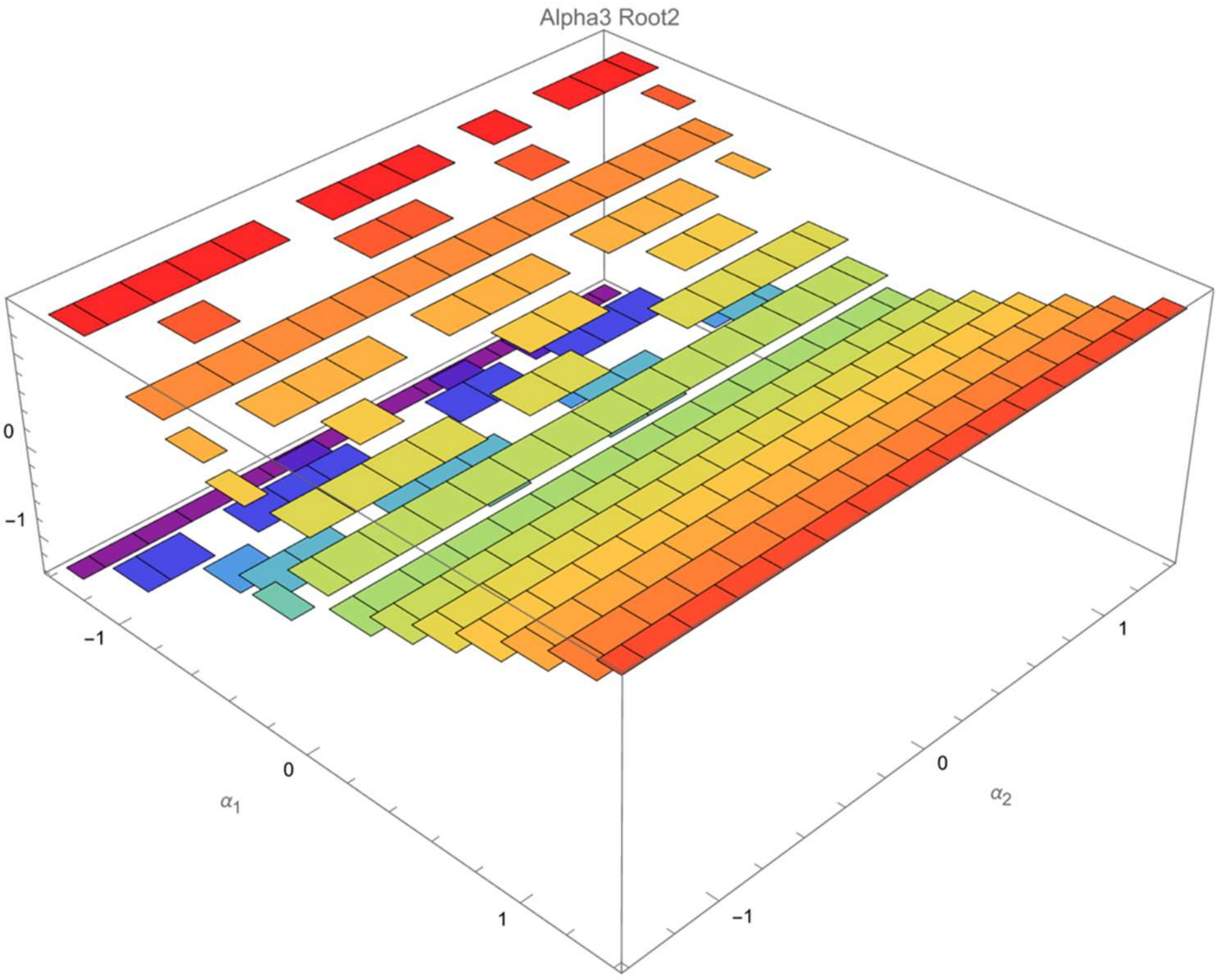

**Figure 5.** The value of $\alpha_3$ in the other roots of ($\alpha_3$,$\alpha_4$).

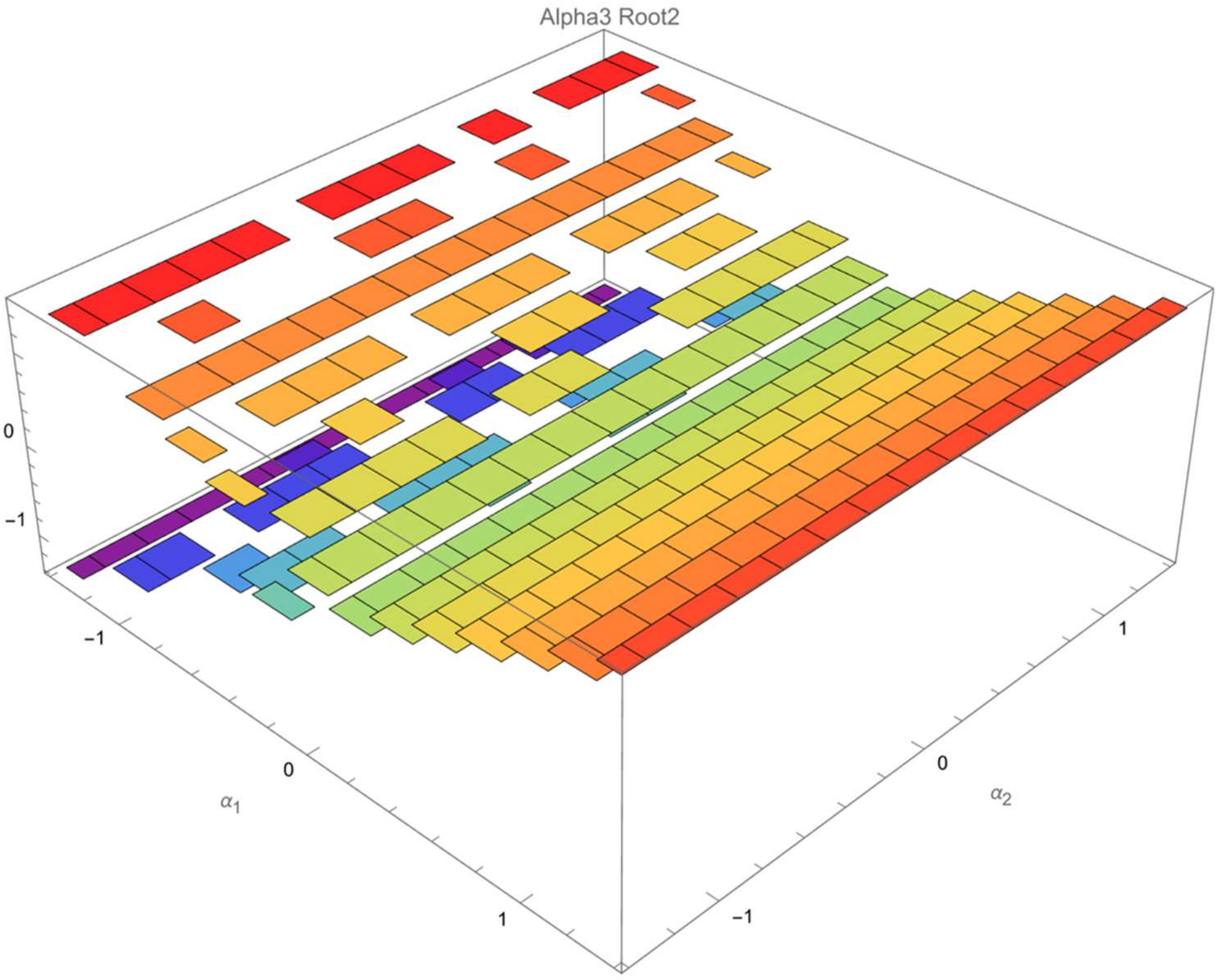

**Figure 6.** The value of $\alpha_4$ in the other root of ($\alpha_3$,$\alpha_4$).

Show two values of $\alpha_3$ in the same figure (Figure 7). Show two values of $\alpha_4$ in the same figure (Figure 8).

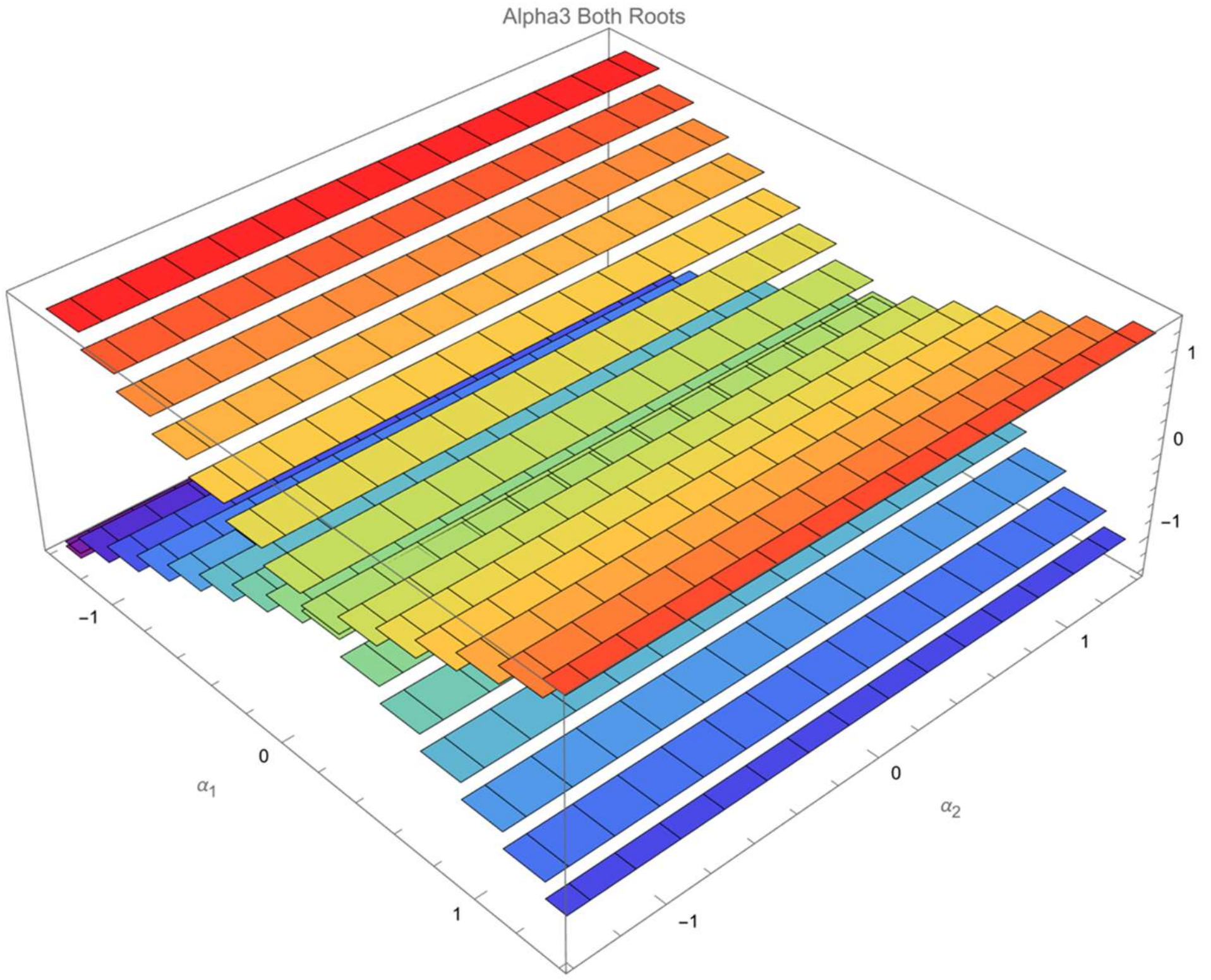

**Figure 7.** The value of $\alpha_3$ in both roots of $(\alpha_3,\alpha_4)$.

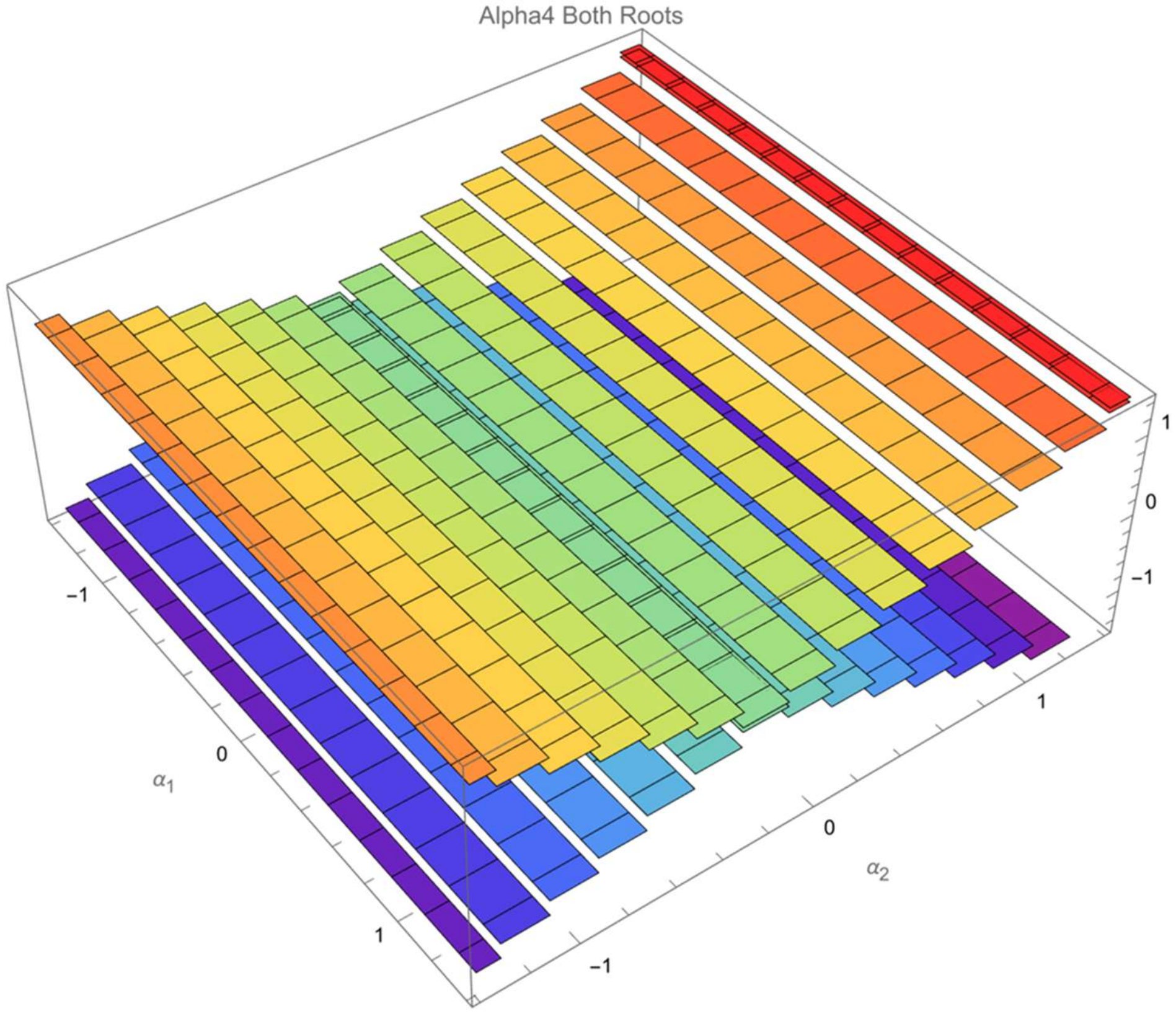

**Figure 8.** The value of $\alpha_4$ in both roots of $(\alpha_3,\alpha_4)$.

Although the regularity of the distribution of $\alpha_3$ in either root of $(\alpha_3,\alpha_4)$, Figure 3 and Figure 5, can hardly be tracked, Figure 7 shows that both $\alpha_3$ lie on the two planes. One plane increases $\alpha_3$ when $\alpha_1$ increase. Define this plane "+" plane. The other plane decreases $\alpha_3$ when $\alpha_1$ increase. Define this plane "−" plane.

Similarly, Figure 8 shows that both $\alpha_4$ also lie on another two planes. One plane increases $\alpha_4$ when $\alpha_2$ increase. Define this plane "+" plane. The other plane decreases $\alpha_4$ when $\alpha_2$ increase. Define this plane "−" plane.

With these definitions, the roots of $(\alpha_3,\alpha_4)$ can be classified by the mark $(\pm,\pm)$. For example, (+,−) represents the root of $(\alpha_3,\alpha_4)$, whose $\alpha_3$ lies on "+" plane and $\alpha_4$ lies on "−" plane.

Observing Figure 3 – Figure 6, there are only 7 categories of the roots of $(\alpha_3,\alpha_4)$. They are (+,+), (−,−), (+,?), (?,+), (−,?), (?,−), (?,?). "?" in either plane represents the $\alpha_3$ or $\alpha_4$ lies in the intersection of the relevant "+" plane and "−" plane.

Note that no roots of $(\alpha_3,\alpha_4)$ belong to (+,−) or (−,+).

These definitions will facilitate the discussions on Two Color Map Theorem, detailed in the next section.

Observing Formula (6), we assert that $R(\alpha_1,\alpha_2,\alpha_3,\alpha_4)$ is continuous, given $\alpha_1$, $\alpha_2$, $\alpha_3$, and $\alpha_4$ are continuous.

**Proposition 1.** *Given $R_\phi(\alpha_1,\alpha_2,\alpha_3,\alpha_4) = 0$ and $R_\theta(\alpha_1,\alpha_2,\alpha_3,\alpha_4) = 0$, the necessary condition to receive an invertible decoupling matrix is*

$$\forall\alpha_1, \alpha_2, \alpha_3, \alpha_4, R(\alpha_1,\alpha_2,\alpha_3,\alpha_4) > 0 \tag{10}$$

*or*

$$\forall\alpha_1, \alpha_2, \alpha_3, \alpha_4, R(\alpha_1,\alpha_2,\alpha_3,\alpha_4) < 0. \tag{11}$$

**Proof of Proposition 1.** Considering that $R(\alpha_1,\alpha_2,\alpha_3,\alpha_4)$ is continuous, the rebuttal method yields this result. □

Based on Proposition 1, the roots of $(\alpha_3,\alpha_4)$ previously found are further classified into the ones receiving positive $R(\alpha_1,\alpha_2,\alpha_3,\alpha_4)$ and the ones receiving negative $R(\alpha_1,\alpha_2,\alpha_3,\alpha_4)$.

Figure 9 plots the roots of $(\alpha_3,\alpha_4)$ in Figure 3 and Figure 4 receiving positive $R(\alpha_1,\alpha_2,\alpha_3,\alpha_4)$ with $\alpha_3$ in red and $\alpha_4$ in blue. The value of the relevant $R(\alpha_1,\alpha_2,\alpha_3,\alpha_4)$ is illustrated in Figure 10.

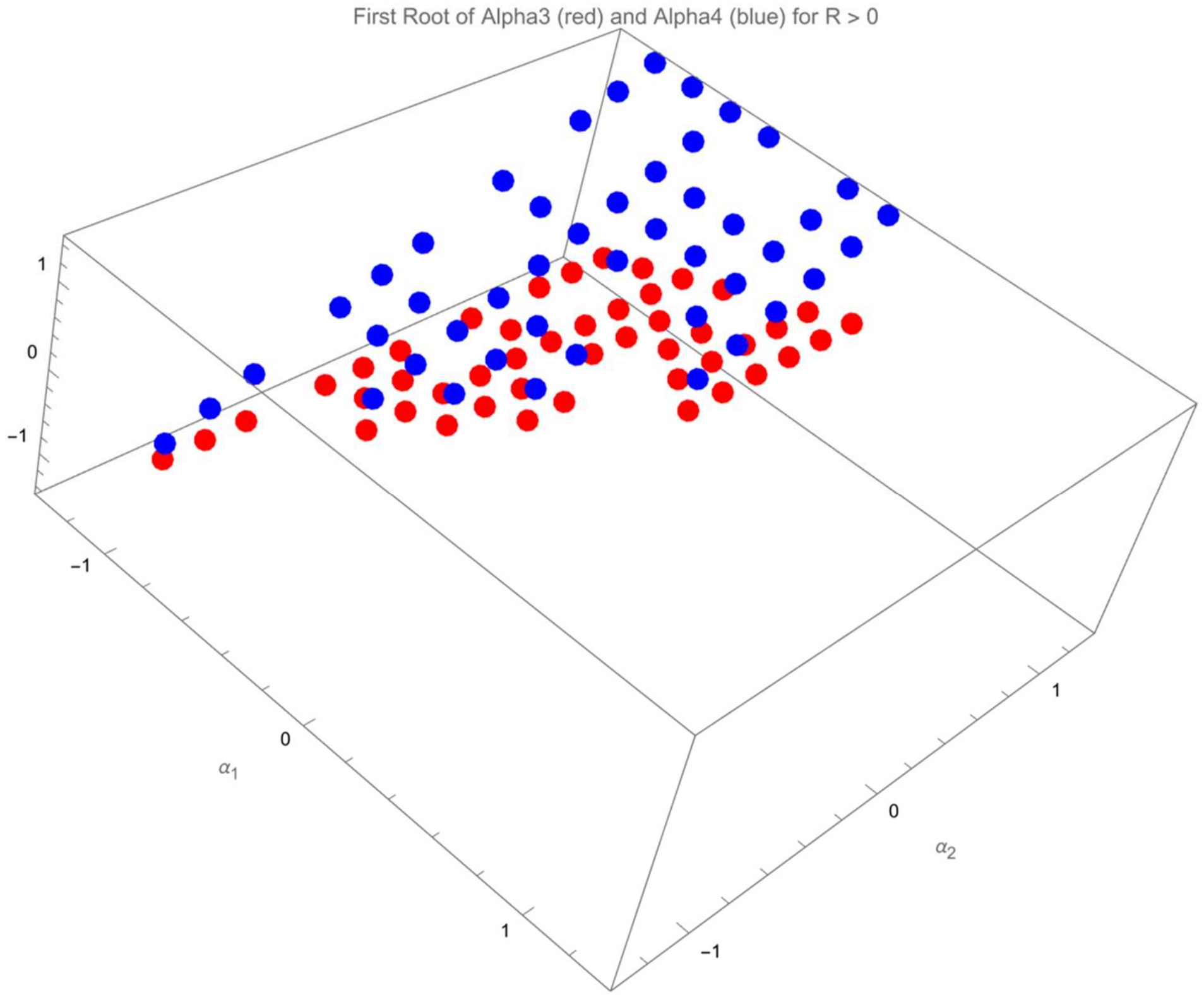

**Figure 9.** The value of $\alpha_3$ and $\alpha_4$ of one of the roots of $(\alpha_3,\alpha_4)$ receiving positive $R(\alpha_1,\alpha_2,\alpha_3,\alpha_4)$. The red points represent $\alpha_3$. The blue points represent $\alpha_4$.

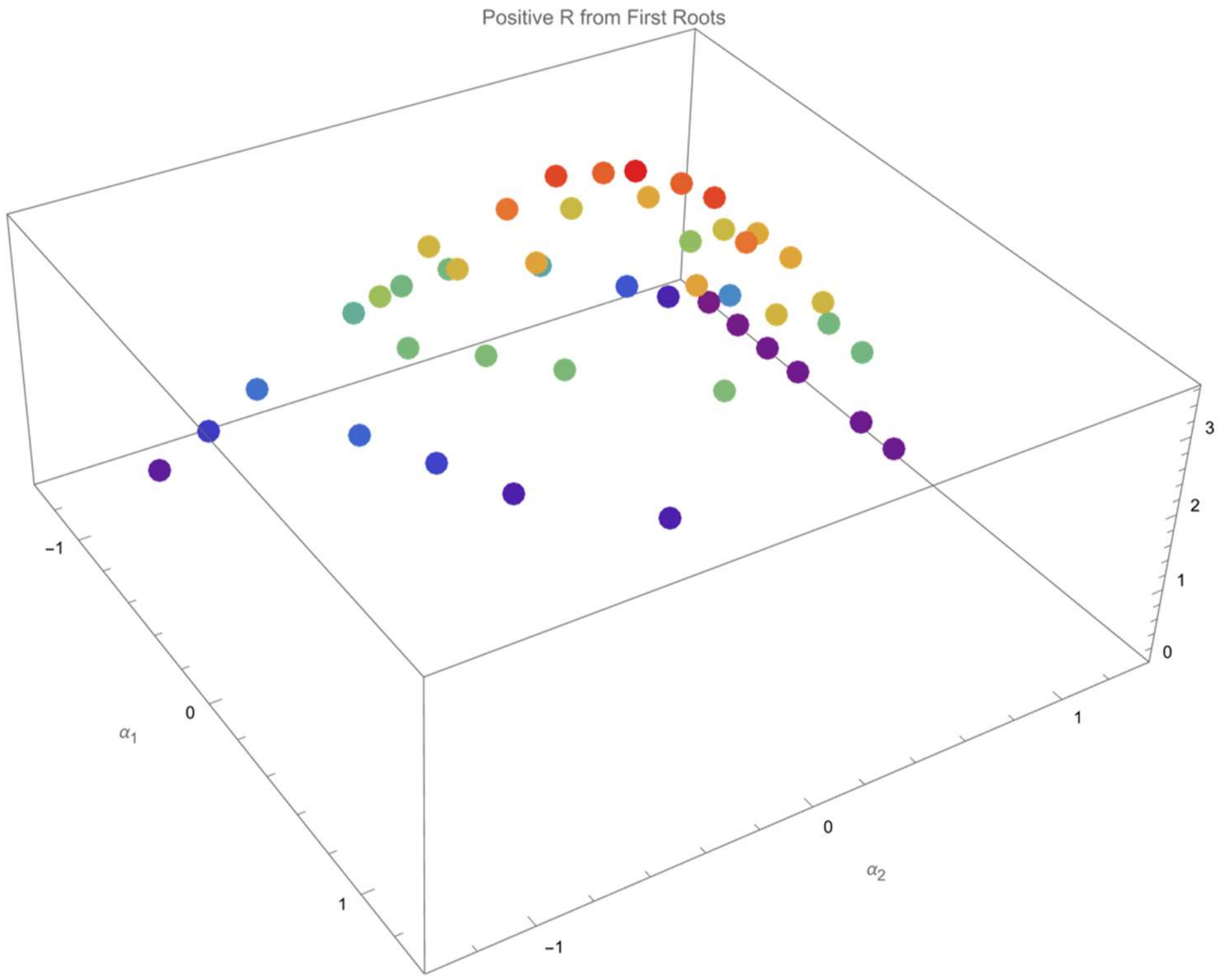

**Figure 10.** One of the roots of $(\alpha_3,\alpha_4)$ receiving positive $R(\alpha_1,\alpha_2,\alpha_3,\alpha_4)$ and its value.

Figure 11 plots the roots of ($\alpha_3$,$\alpha_4$) in Figure 5 and Figure 6 receiving positive $R(\alpha_1,\alpha_2,\alpha_3,\alpha_4)$ with $\alpha_3$ in red and $\alpha_4$ in blue. The value of the relevant $R(\alpha_1,\alpha_2,\alpha_3,\alpha_4)$ is illustrated in Figure 12.

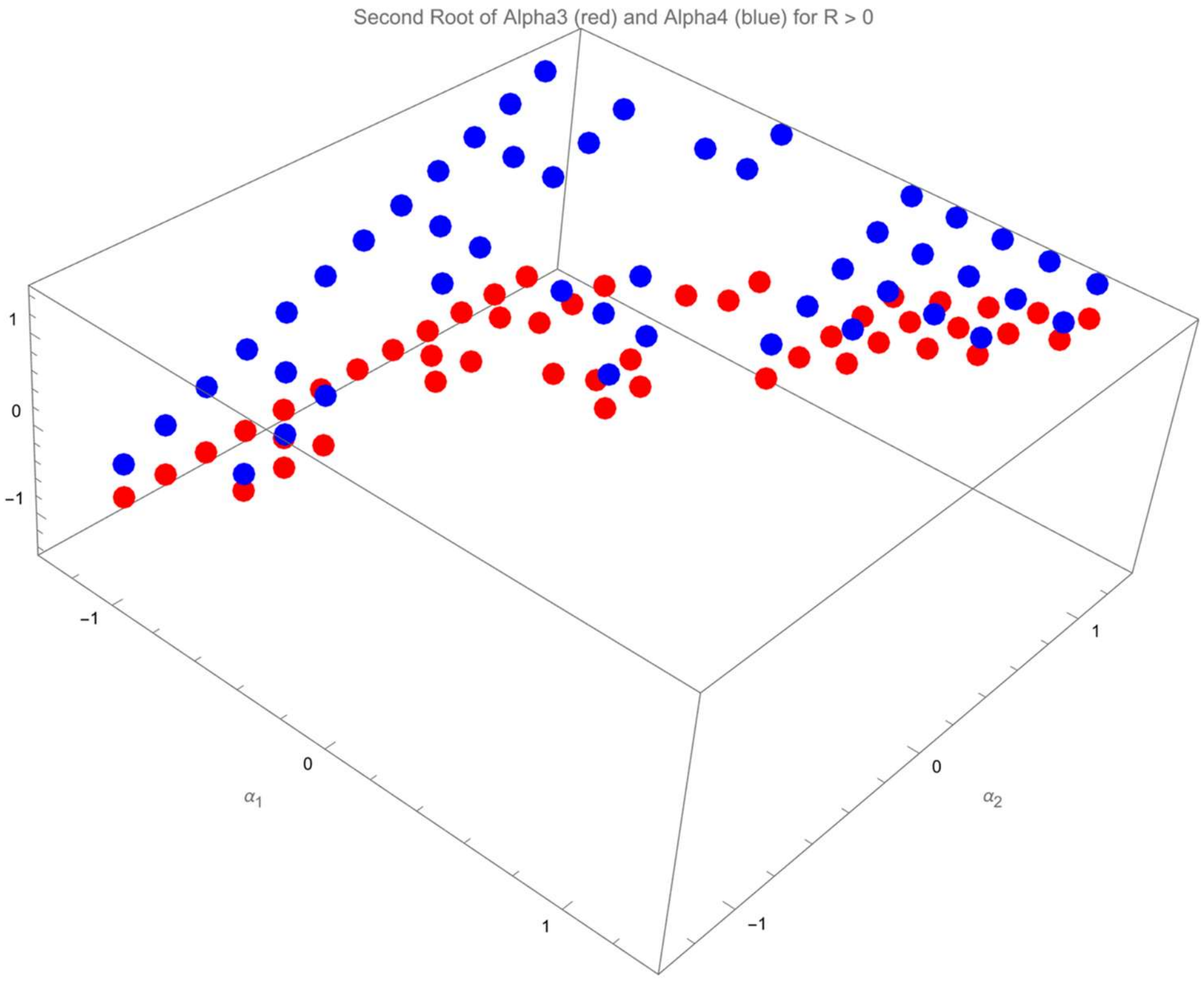

**Figure 11.** The value of $\alpha_3$ and $\alpha_4$ of the other root of ($\alpha_3$,$\alpha_4$) receiving positive $R(\alpha_1,\alpha_2,\alpha_3,\alpha_4)$. The red points represent $\alpha_3$. The blue points represent $\alpha_4$.

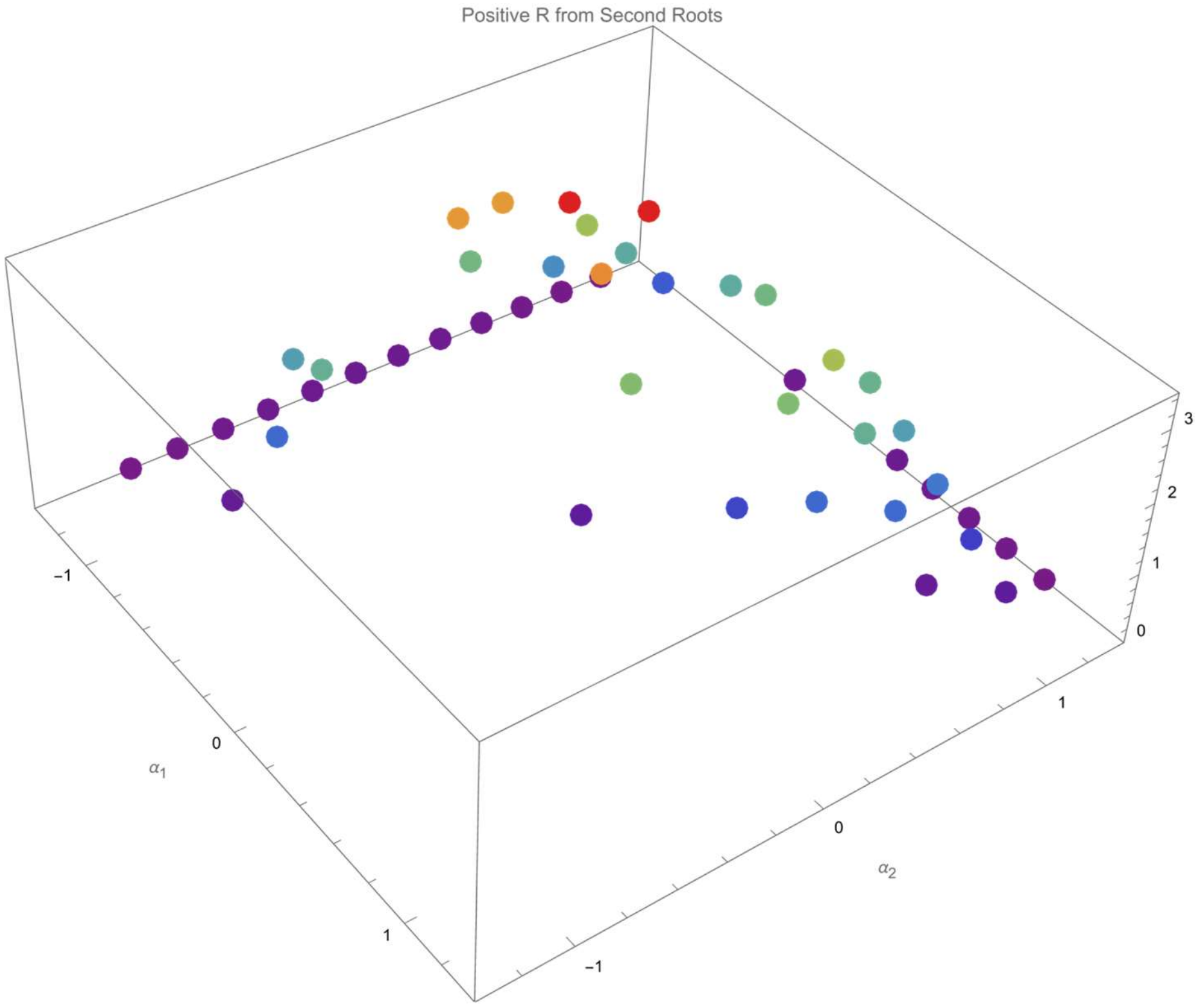

**Figure 12.** The other root of $(\alpha_3,\alpha_4)$ receiving positive $R(\alpha_1,\alpha_2,\alpha_3,\alpha_4)$ and its value.

Note that both roots of $(\alpha_3,\alpha_4)$ receiving positive $R(\alpha_1,\alpha_2,\alpha_3,\alpha_4)$, Figure 9 and Figure 11, belong to (+,+) (+,?), and (?,+) types. Further, the projection of these applicable $(\alpha_3,\alpha_4)$ in both roots fully occupies triangular area of $(\alpha_1,\alpha_2)$ without overlapping, indicating that there is only one root of $(\alpha_3,\alpha_4)$ meeting $R(\alpha_1,\alpha_2,\alpha_3,\alpha_4) > 0$ for each $(\alpha_1,\alpha_2)$ in the triangular area. This triangular area is governed by three vertices: $(\alpha_1,\alpha_2) = (-7\pi/16, 7\pi/16)$, $(\alpha_1,\alpha_2) = (-7\pi/16, -5\pi/16)$, $(\alpha_1,\alpha_2) = (5\pi/16, 7\pi/16)$.

Similarly, the distributions of the both roots of $(\alpha_3,\alpha_4)$ receiving negative $R(\alpha_1,\alpha_2,\alpha_3,\alpha_4)$ are plotted in Figure 13 and Figure 14 with $\alpha_3$ in red and $\alpha_4$ in blue.

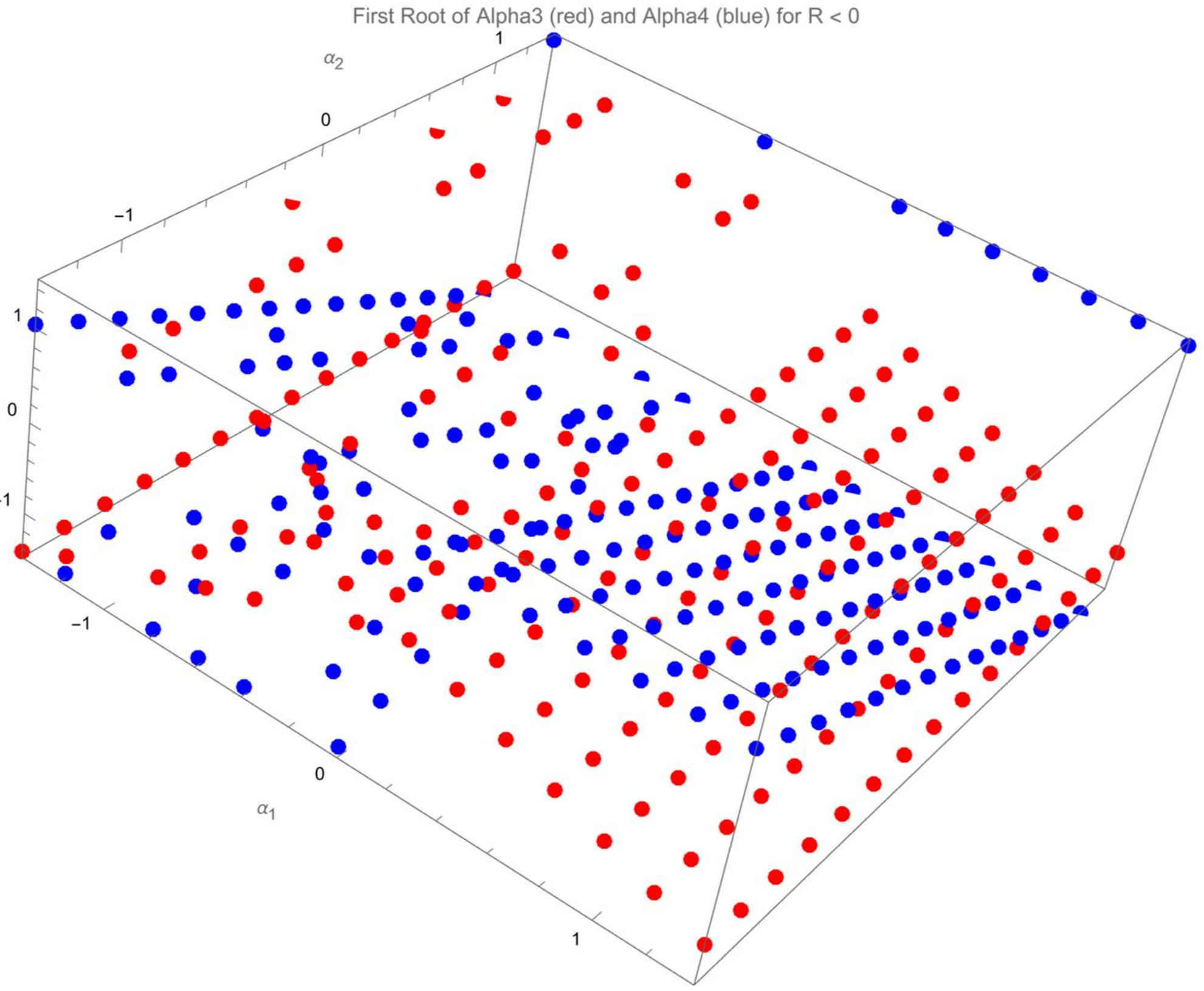

**Figure 13.** The value of $\alpha_3$ and $\alpha_4$ of one of the roots of $(\alpha_3,\alpha_4)$ receiving negative $R(\alpha_1,\alpha_2,\alpha_3,\alpha_4)$. The red points represent $\alpha_3$. The blue points represent $\alpha_4$.

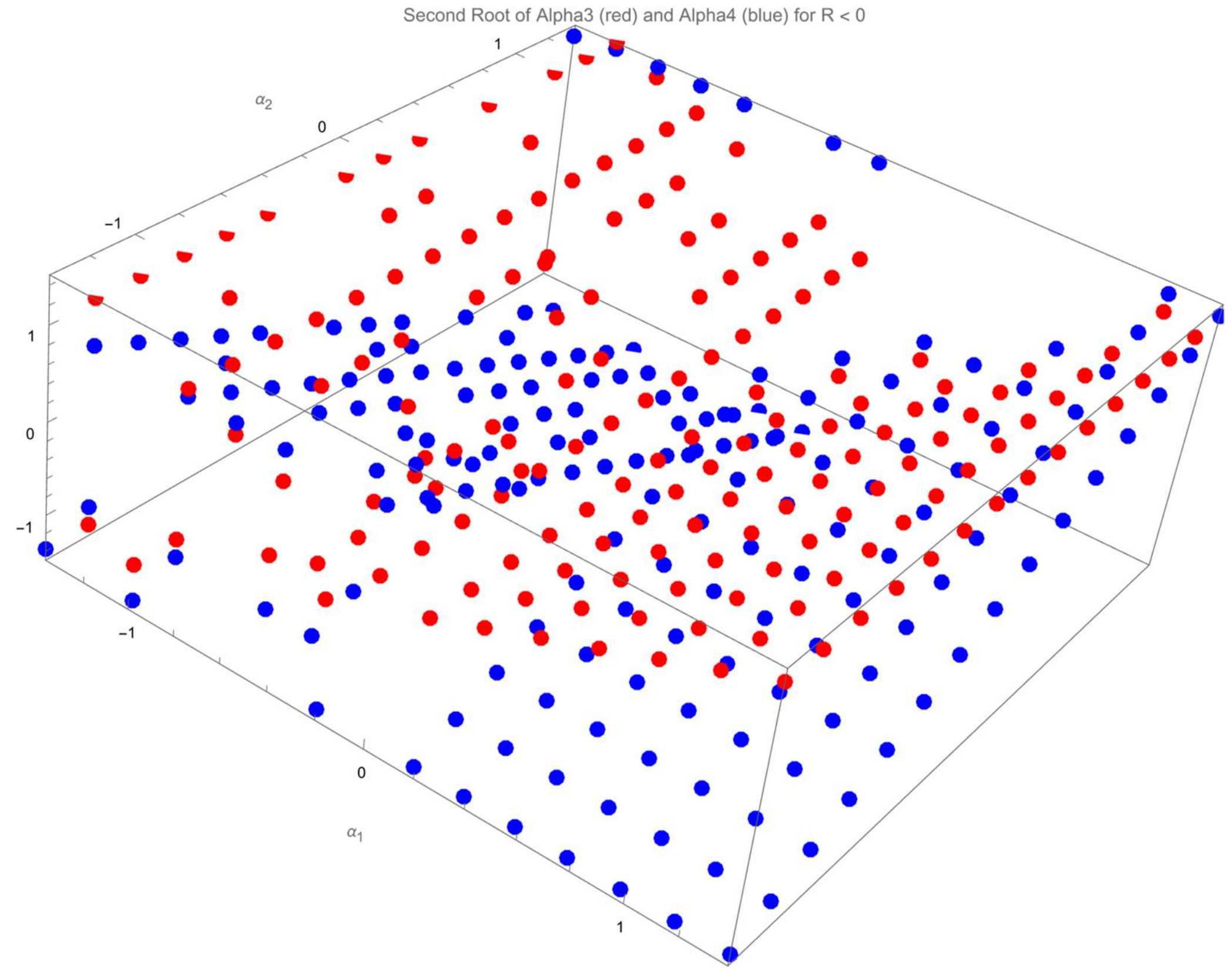

**Figure 14.** The value of $\alpha_3$ and $\alpha_4$ of the other root of $(\alpha_3,\alpha_4)$ receiving negative $R(\alpha_1,\alpha_2,\alpha_3,\alpha_4)$. The red points represent $\alpha_3$. The blue points represent $\alpha_4$.

Both roots of $(\alpha_3,\alpha_4)$ receiving negative $R(\alpha_1,\alpha_2,\alpha_3,\alpha_4)$, Figure 13 and Figure 14, covers all 7 types: (+,+), (−,−), (+,?), (?,+), (−,?), (?,−), (?,?).

In addition, the projection of these applicable $(\alpha_3,\alpha_4)$ in both roots fully occupies triangular area of $(\alpha_1,\alpha_2)$ governed by three vertices: $(\alpha_1,\alpha_2) = (-3\pi/8,-3\pi/8)$, $(\alpha_1,\alpha_2) = (3\pi/8,3\pi/8)$, $(\alpha_1,\alpha_2) = (3\pi/8,-3\pi/8)$ with overlapping at some $(\alpha_1,\alpha_2)$, indicating that there are two roots of $(\alpha_3,\alpha_4)$ meeting $R(\alpha_1,\alpha_2,\alpha_3,\alpha_4) < 0$ for some $(\alpha_1,\alpha_2)$ in this triangular area. In other words, the number of the roots of $(\alpha_3,\alpha_4)$ receiving negative $R(\alpha_1,\alpha_2,\alpha_3,\alpha_4)$ in this region is 1 or 2.

So far, all the $(\alpha_1,\alpha_2,\alpha_3,\alpha_4)$ satisfying the necessary condition to receive an invertible gait, Formula (7) – (9), on the four-dimensional gait surface have been classified.

One may plan a gait by varying $(\alpha_1,\alpha_2)$ continuously and recording the corresponding $(\alpha_3,\alpha_4)$. The resulting $(\alpha_1,\alpha_2,\alpha_3,\alpha_4)$ is then taken as a gait. This pattern works in the case $R(\alpha_1,\alpha_2,\alpha_3,\alpha_4) > 0$.

While two obstacles may hinder the application of this pattern when $R(\alpha_1,\alpha_2,\alpha_3,\alpha_4) < 0$. Firstly, there are the region of $(\alpha_1,\alpha_2)$ returning two roots to $(\alpha_3,\alpha_4)$ when $R(\alpha_1,\alpha_2,\alpha_3,\alpha_4) < 0$; determining the proper root, $(\alpha_3,\alpha_4)$, for these $(\alpha_1,\alpha_2)$ is necessary. Secondly, varying $(\alpha_1,\alpha_2)$ continuously does not necessarily guarantee that $(\alpha_3,\alpha_4)$ is varying continuously. Ironically, one of the advantages of the gait plan is the continuous change in the tilting -angles.

With these concerns, a theory helping select the proper $(\alpha_3,\alpha_4)$ for every given $(\alpha_1,\alpha_2)$ is demanded to receive a continuous gait when $R(\alpha_1,\alpha_2,\alpha_3,\alpha_4) < 0$. Two Color Map Theorem is consequently introduced.

## 4. Two Color Map Theorem

The continuous gait is guaranteed by a sophisticated $(\alpha_3,\alpha_4)$-picking theorem, Two Color Map Theorem. It is an augmented rule for varying $(\alpha_1,\alpha_2,\alpha_3,\alpha_4)$ while $R(\alpha_1,\alpha_2,\alpha_3,\alpha_4) < 0$. Without specification, the region in the rest of this section represents the region of $(\alpha_1,\alpha_2)$ with negative $R(\alpha_1,\alpha_2,\alpha_3,\alpha_4)$.

Recalling $(\alpha_3,\alpha_4)$ receiving $R(\alpha_1,\alpha_2,\alpha_3,\alpha_4) < 0$, there are altogether 7 types of $(\alpha_3,\alpha_4)$ in this case: (+,+), (−,−), (+,?), (?,+), (−,?), (?,−), (?,?).

Notice that "+" plane and "−" plane of $\alpha_3$ intersects at "?", which is a line. Also, "+" plane and "−" plane of $\alpha_4$ intersects at a line marked by "?". $\alpha_3$ is not allowed to move from "+" to "−" or to move from "−" to "+". Both cases will introduce the discontinuous change of $\alpha_3$. Similarly, $\alpha_4$ is not allowed to move from "+" to "−" or to move from "−" to "+". Both cases will introduce the discontinuous change of $\alpha_4$.

To change the plane, the points on the intersecting lines "?" of the two planes, "+" to "−", is required to be an intermediate state. For example, the relevant changes from "+" to "?" to "−" and from "−" to "?" to "+" are both continuous.

The allowed types for two adjacent $\alpha_3$ or $\alpha_4$ are lined in Figure 15 (a). The allowed types for two adjacent $(\alpha_3,\alpha_4)$ are lined in Figure 15 (b) where the case (?,?) is omitted since it accommodates all types of adjacent roots.

The allowed adjacent roots and root pairs.

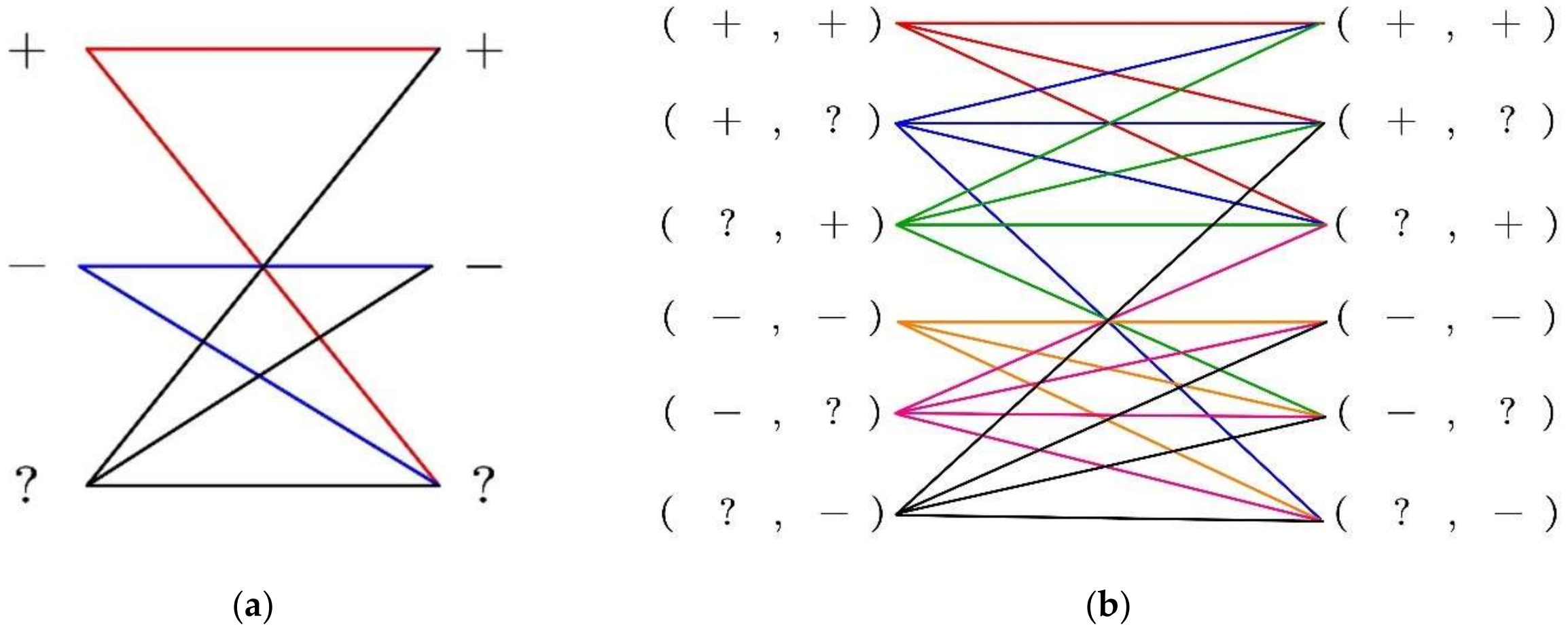

**Figure 15.** The allowed adjacent roots are lined: (**a**) The allowed types of the adjacent $\alpha_3$ or $\alpha_4$; (**b**) The allowed types of the adjacent $(\alpha_3,\alpha_4)$. The case (?,?) is not plotted.

Paint $(\alpha_1,\alpha_2)$ red if the corresponding $(\alpha_3,\alpha_4)$ belongs to (+,+). Paint $(\alpha_1,\alpha_2)$ blue if the corresponding $(\alpha_3,\alpha_4)$ belongs to (−,−). Paint $(\alpha_1,\alpha_2)$ half red half blue if there are two corresponding $(\alpha_3,\alpha_4)$ which belong to (−,−) and (+,+), respectively. Marking the rest categories directly, Figure 16 paints the whole region of $(\alpha_1,\alpha_2)$ of interest ($R(\alpha_1,\alpha_2,\alpha_3,\alpha_4) < 0$).

It can be seen that most $(\alpha_1,\alpha_2)$ receive two applicable $(\alpha_3,\alpha_4)$, belonging to (+,+) and (−,−), respectively. While $(\alpha_1,\alpha_2)$ in the triangular region on the top receive one applicable $(\alpha_3,\alpha_4)$ only, the type of which is (−,−).

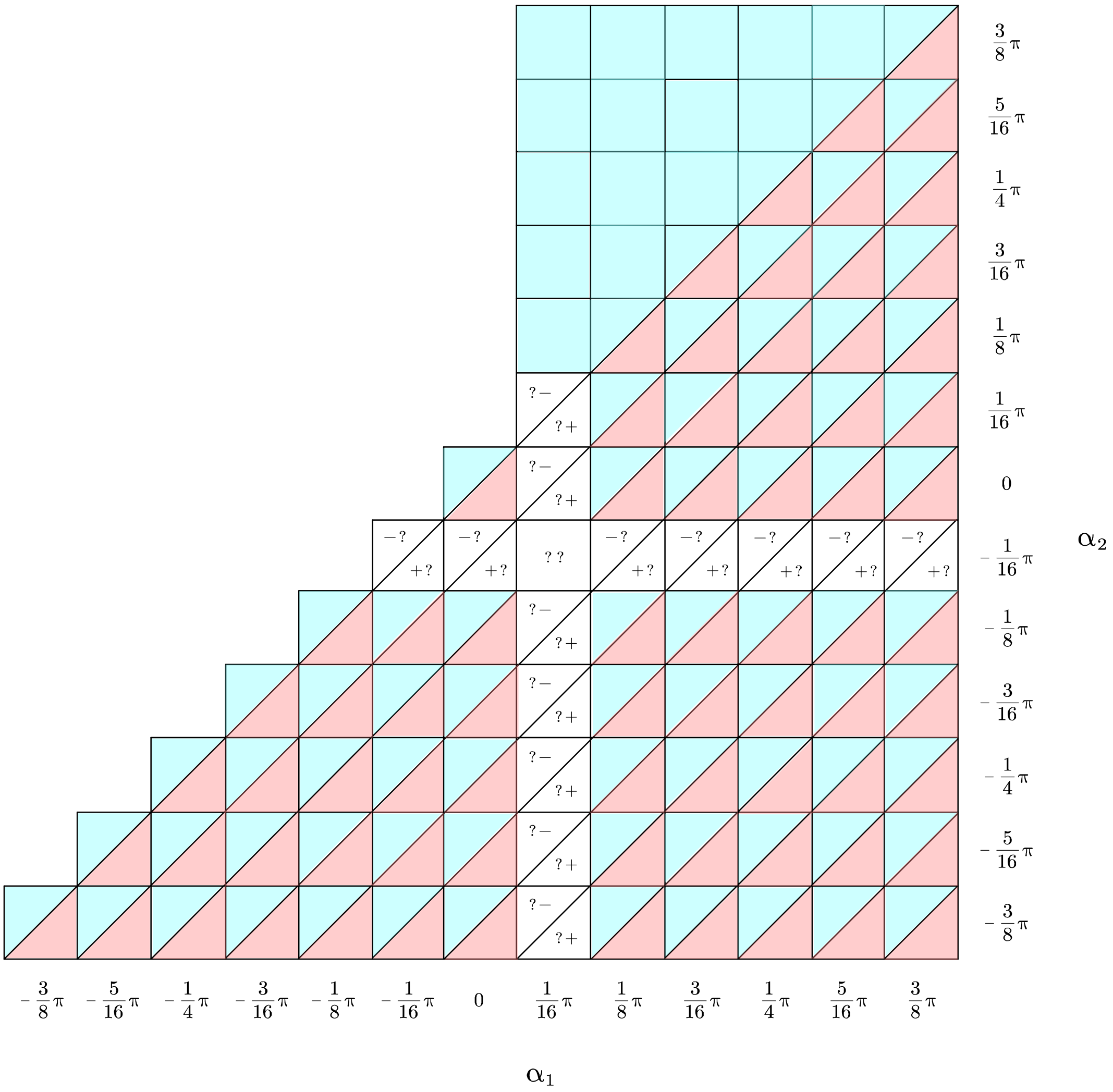

**Figure 16.** The types of the corresponding $(\alpha_3,\alpha_4)$, resulting in $R(\alpha_1,\alpha_2,\alpha_3,\alpha_4) < 0$, on $(\alpha_1,\alpha_2)$ plane. $(\alpha_1,\alpha_2)$ receiving $(\alpha_3,\alpha_4)$ of type (+,+) are painted in red. $(\alpha_1,\alpha_2)$ receiving $(\alpha_3,\alpha_4)$ of type (−,−) are painted in blue. $(\alpha_1,\alpha_2)$ receiving two $(\alpha_3,\alpha_4)$ of both types are painted in half red half blue. The rest $(\alpha_1,\alpha_2)$ are marked based on the types of the corresponding $(\alpha_3,\alpha_4)$, some of which have 2 roots.

As seen in Figure 16, some $(\alpha_1,\alpha_2)$ receive 2 roots of $(\alpha_3,\alpha_4)$. One can pick either root as wish, which can be interpreted as a color decision: decide blue or red for the $(\alpha_1,\alpha_2)$ painted in half red half blue in Figure 16.

The gait plan problem is then equivalent to the following processes: 1. Find an enclosed curve (including a point or an overlapping curve with zero size) in $(\alpha_1,\alpha_2)$ plane in Figure 16. 2. Decide the color for $(\alpha_1,\alpha_2)$ (or decide $(\alpha_3,\alpha_4)$) on the enclosed curve found. 3. Specify the decided four-dimensional curve with direction and time.

The following Two Color Map Theorem restricts the selection of $(\alpha_3,\alpha_4)$, leading to a continuous gait.

**Two Color Map Theorem.** *The planned gait is continuous if the color selections of $(\alpha_1,\alpha_2)$ on the enclosed curve meet the following requirement: The adjacent $(\alpha_1,\alpha_2)$ are in the same color or do not violate the rule in Figure 15.*

**Proof of Two Color Map Theorem.** The same colored adjacent $(\alpha_1,\alpha_2)$ indicate that $\alpha_3$ and $\alpha_4$ will not change the type of the plane when $(\alpha_1,\alpha_2)$ varies. Further, since $(\alpha_1,\alpha_2)$ varies continuously, $(\alpha_3,\alpha_4)$ is consequently continuous. The planned gait $(\alpha_1,\alpha_2,\alpha_3,\alpha_4)$ is a continuous gait. □

So far, the method of planning a continuous gait robust to the attitude change in a tilt-rotor has been elucidated. Several examples are given in the next section.

## 5. Analysis on Typical Acceptable Gaits

This section evaluates four different gaits. Three of them satisfy $R(\alpha_1,\alpha_2,\alpha_3,\alpha_4) < 0$ for any given time. The rest one satisfies $R(\alpha_1,\alpha_2,\alpha_3,\alpha_4) > 0$ for any given time. In comparison, the relevant biased gaits (by scaling) are also evaluated.

The gaits, satisfying $R(\alpha_1,\alpha_2,\alpha_3,\alpha_4) < 0$, analyzed in this section are plotted in Figure 17. Based on this choice, the color of all $(\alpha_1,\alpha_2)$ travelled by the enclosed curve, Gait 1 in Figure 17, shall be painted blue based on Two Color Map Theorem.

On the other hand, the color of $(\alpha_1,\alpha_2)$ travelled by the enclosed curve, Gait 2 and Gait 3 in Figure 17, can be painted either all in blue or all in red based on Two Color Map Theorem.

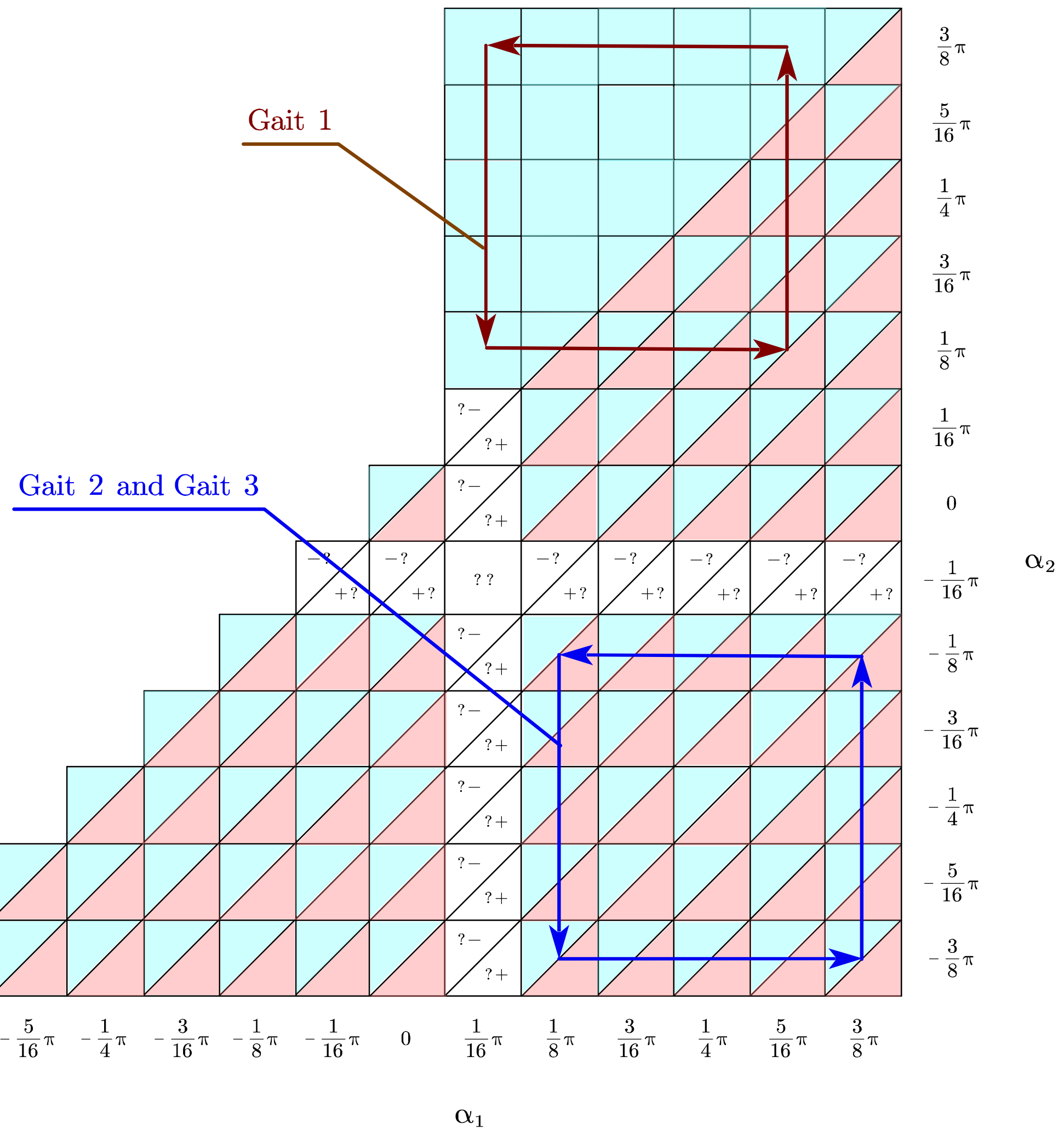

**Figure 17.** The projections of all three designed gaits satisfying $R(\alpha_1,\alpha_2,\alpha_3,\alpha_4) < 0$ are the enclosed curves. The colors of $(\alpha_1,\alpha_2)$ travelled by each gait should satisfy Two Color Map Theorem.

As for the gait (Gait 4, not sketched) satisfying $R(\alpha_1,\alpha_2,\alpha_3,\alpha_4) > 0$, there is only one root of $(\alpha_3,\alpha_4)$, belonging to (+,+), corresponding to a given $(\alpha_1,\alpha_2)$ (see Figure 9 and Figure 11). Varying $(\alpha_1,\alpha_2)$ naturally guarantees a continuous change in $(\alpha_1,\alpha_2,\alpha_3,\alpha_4)$.

The robustness to the disturbance in roll angle and pitch angles of each proposed gait is evaluated. The unacceptable attitudes (roll and pitch) for each gait is found by violating Formula (1), e.g., equal the left side of Formula (1) zero.

The biased gaits not on the four-dimensional gait surface are coined to compare the robustness with the gaits on the four-dimensional gait surface. These biased gaits are generated by maintaining $(\alpha_1,\alpha_2)$ identical to the acceptable gaits while scaling $(\alpha_3,\alpha_4)$ throughout the entire gait:

$$\alpha_1 \leftarrow \alpha_1, \tag{12}$$

$$\alpha_2 \leftarrow \alpha_2, \tag{13}$$

$$\alpha_3 \leftarrow \eta \cdot \alpha_3, \tag{14}$$

$$\alpha_4 \leftarrow \eta \cdot \alpha_4, \tag{15}$$

where $\eta \in (0,1)$ is the scaling coefficient, which will be specified in each gait.

Note that though scaling evenly in each tilting-angle has been proved a valid approach in modifying a gait receiving a singular decoupling matrix [27], the discussions on scaling unevenly, e.g., scaling $\alpha_3$ and $\alpha_4$ only, have not been addressed yet.

The period of the gait ($T$) is set as 1 second.

Define the $(\alpha_1,\alpha_2,\alpha_3,\alpha_4)$ as a vertex of a four-dimensional curve if and only if

$$\alpha_1 \in \left\{\min_{t\in T}(\alpha_1),\max_{t\in T}(\alpha_1)\right\}, \tag{16}$$

$$\alpha_2 \in \left\{\min_{t\in T}(\alpha_2),\max_{t\in T}(\alpha_2)\right\}, \tag{17}$$

$$\alpha_3 \in \left\{\min_{t\in T}(\alpha_3),\max_{t\in T}(\alpha_3)\right\}, \tag{18}$$

$$\alpha_4 \in \left\{\min_{t\in T}(\alpha_4),\max_{t\in T}(\alpha_4)\right\}. \tag{19}$$

The number of the vertices of the four-dimensional curve in each designed gait in this research is four.

### *4.1. Gait 1*

All $(\alpha_1,\alpha_2)$ projected by the four-dimensional curve (Gait 1) are painted in blue to meet Two Color Map Theorem.

The vertices of the gait $(\alpha_1,\alpha_2,\alpha_3,\alpha_4)$ are: $(5\pi/16,\pi/8,-0.648,-0.727)$, $(5\pi/16,3\pi/8,-0.648,-1.512)$, $(\pi/16,3\pi/8,0.138,-1.512)$, $(\pi/16,\pi/8,0.138,-0.727)$.

In comparison, the corresponding biased gait for Gait 1 is coined by setting the scaling coefficient, $\eta$ in Formula (14) and (15), as 80%.

Figure 18 plots the unacceptable attitudes for the gait on the four-dimensional gait surface (red curves) and for the biased gait (blue curves).

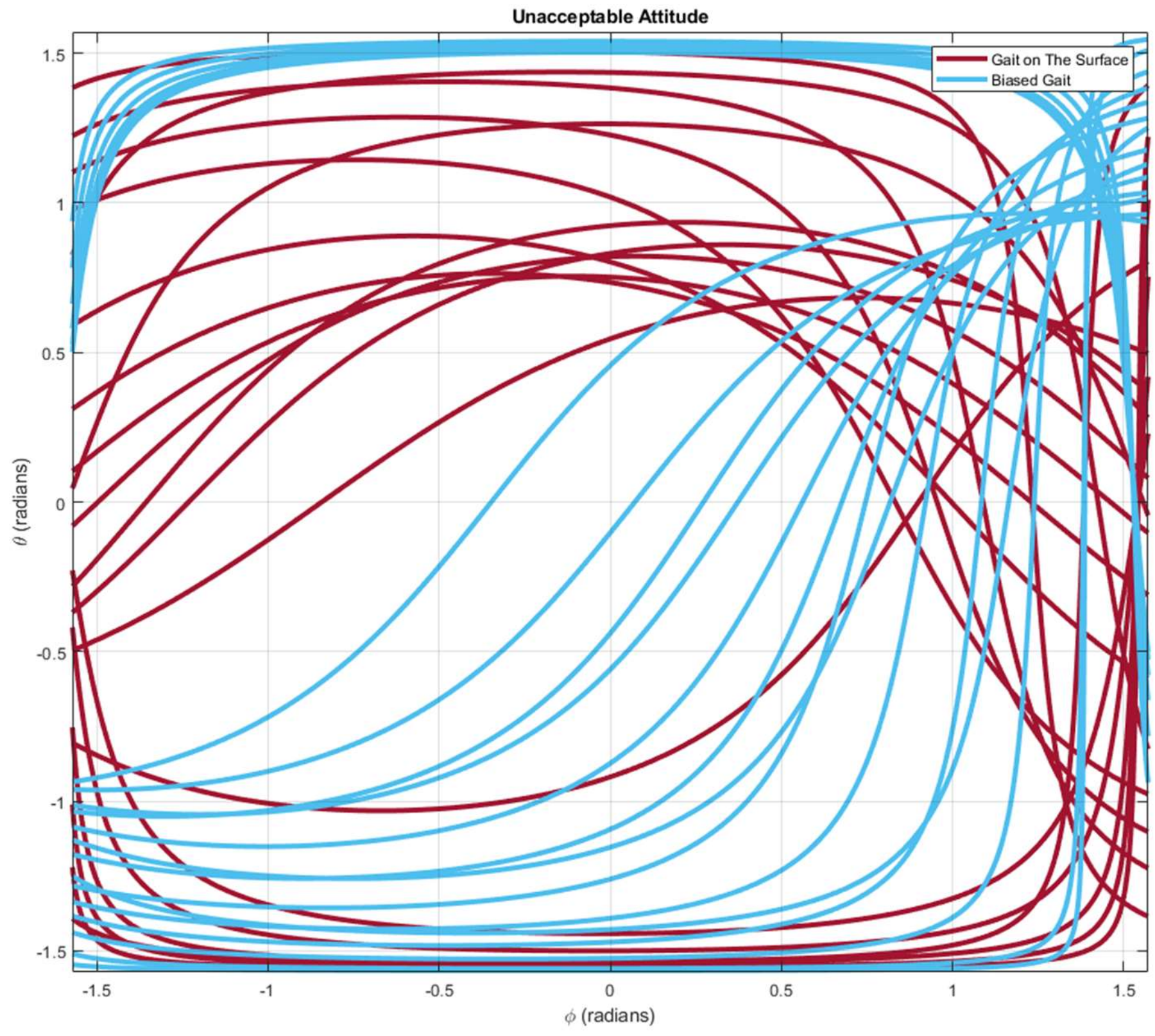

**Figure 18.** The unacceptable attitudes for Gait 1 and for biased Gait 1. The red curve represents the attitudes receiving the singular decoupling matrix in Gait 1. The blue curve represents the attitudes receiving the singular decoupling matrix in biased Gait 1.

It can be seen that the red curves are generally further to the origin, $(\phi,\theta) = (0,0)$, comparing with the blue curves, demonstrating that the gait on the four-dimensional gait surface is allowed to move in a wilder region of attitude.

*4.2. Gait 2 and 3*

All $(\alpha_1,\alpha_2)$ projected by the four-dimensional curve (Gait 2 and Gait 3) can either be painted in blue or in red to meet Two Color Map Theorem. This leads to Gait 2 (all in blue) and Gait 3 (all in red).

The vertices of the tilting-angles $(\alpha_1,\alpha_2,\alpha_3,\alpha_4)$ in Gait 2 are: $(3\pi/8,-3\pi/8,-0.844,0.844)$, $(3\pi/8,-\pi/8,-0.844,0.059)$, $(\pi/8,-\pi/8,-0.059,0.059)$, $(\pi/8,-3\pi/8,-0.059,0.844)$. The corresponding biased gait for Gait 2 is coined by setting the scaling coefficient, $\eta$ in Formula (14) and (15), as 99%.

Figure 19 plots the unacceptable attitudes for the gait on the four-dimensional gait surface (red curves) and for the biased gait (blue curves).

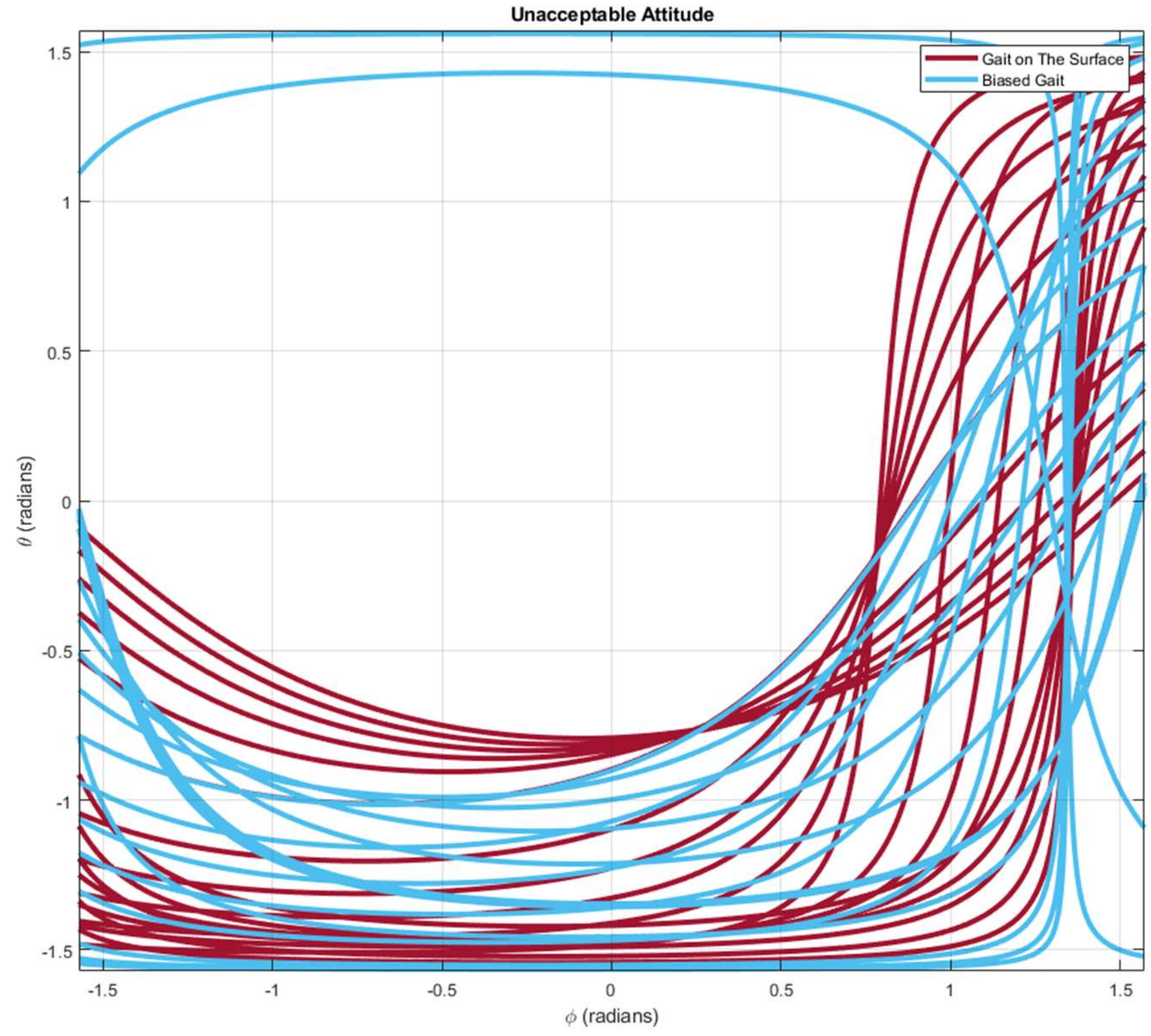

**Figure 19.** The unacceptable attitudes for Gait 2 and for biased Gait 2. The red curve represents the attitudes receiving the singular decoupling matrix in Gait 2. The blue curve represents the attitudes receiving the singular decoupling matrix in biased Gait 2.

Similarly, the red curves are generally further to the origin, $(\phi,\theta) = (0,0)$, comparing with the blue curves, demonstrating that the gait on the four-dimensional gait surface is allowed to move in a wilder region of attitude.

The vertices of the tilting-angles $(\alpha_1,\alpha_2,\alpha_3,\alpha_4)$ in Gait 3 are: $(3\pi/8,-3\pi/8,1.178,-1.178)$ , $(3\pi/8,-\pi/8,1.178,-0.393)$ , $(\pi/8,-\pi/8,0.393,-0.393)$ , $(\pi/8,-3\pi/8,0.393,-1.178)$. The corresponding biased gait for Gait 3 is coined by setting the scaling coefficient, $\eta$ in Formula (14) and (15), as 80%.

Figure 20 plots the unacceptable attitudes for the gait on the four-dimensional gait surface (red curves) and for the biased gait (blue curves).

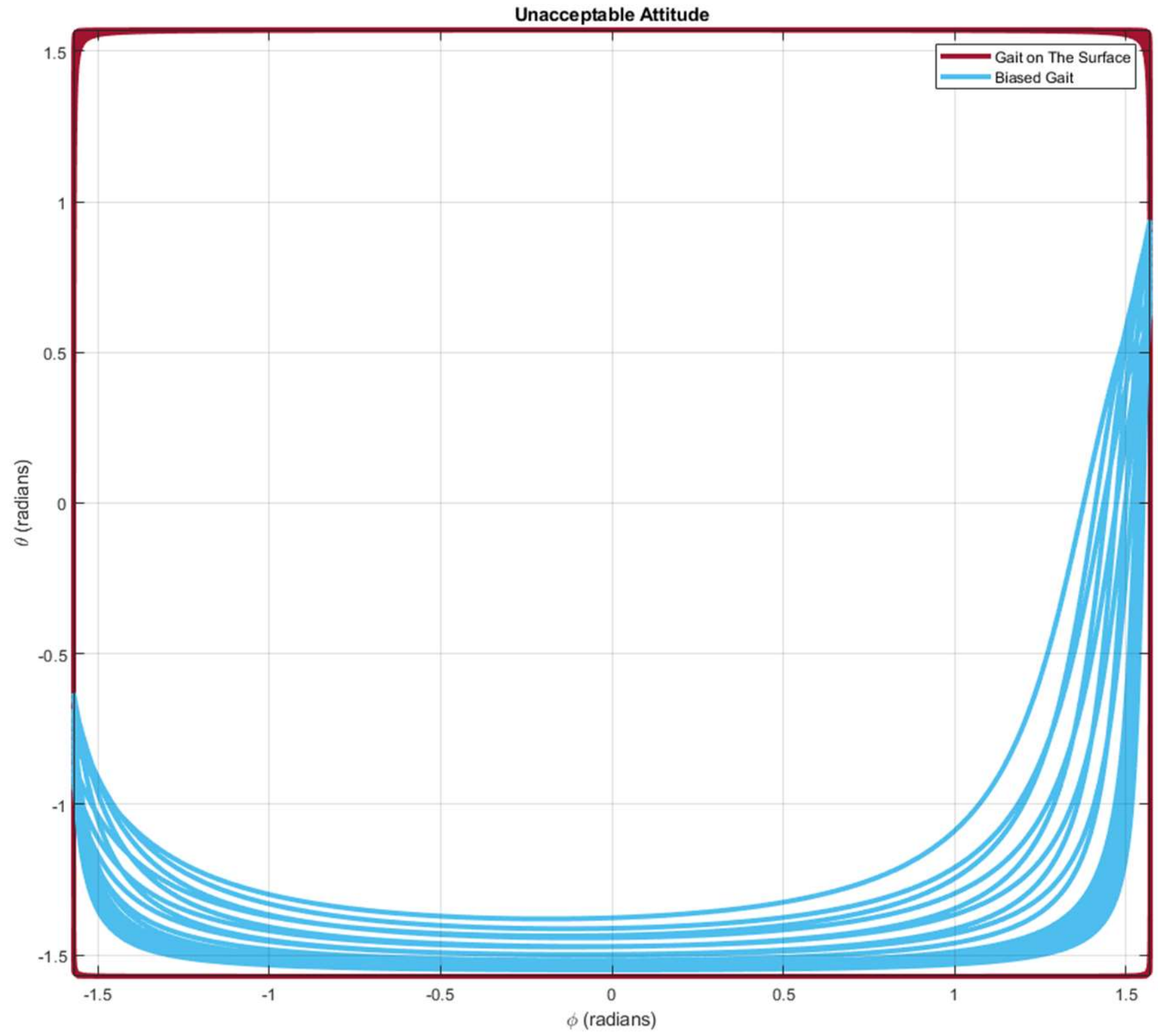

**Figure 20.** The unacceptable attitudes for Gait 3 and for biased Gait 3. The red curve represents the attitudes receiving the singular decoupling matrix in Gait 3. The blue curve represents the attitudes receiving the singular decoupling matrix in biased Gait 3.

Clearly, the red curves are generally further to the origin, $(\phi,\theta) = (0,0)$, comparing with the blue curves, demonstrating that the gait on the four-dimensional gait surface is allowed to move in a wilder region of attitude. The region of admissible attitudes enlarges nearly to the whole region of $\phi \in (-\pi/2,\pi/2), \theta \in (-\pi/2,\pi/2)$.

*4.3. Gait 4*

Gait 4 is designed on the four-dimensional gait surface satisfying $R(\alpha_1,\alpha_2,\alpha_3,\alpha_4) > 0$ (Figure 9 and Figure 11). Only one root of $(\alpha_3,\alpha_4)$ exists for a given $(\alpha_1,\alpha_2)$ in this region, the triangular area is governed by three vertices: $(\alpha_1,\alpha_2) = (-7\pi/16,7\pi/16)$, $(\alpha_1,\alpha_2) = (-7\pi/16,-5\pi/16)$, $(\alpha_1,\alpha_2) = (5\pi/16,7\pi/16)$.

Moreover, $(\alpha_3,\alpha_4)$ varies continuously if $(\alpha_1,\alpha_2)$ varies continuously; all $(\alpha_1,\alpha_2)$ in this region satisfying $R(\alpha_1,\alpha_2,\alpha_3,\alpha_4) > 0$ belong to $(+,+)$.

The vertices of the tilting-angles $(\alpha_1,\alpha_2,\alpha_3,\alpha_4)$ in Gait 4 are: $(-\pi/8,\pi/8,-0.393,0.393)$, $(-\pi/8,3\pi/8,-0.393,1.178)$, $(-3\pi/8,3\pi/8,-1.178,1.178)$, $(-3\pi/8,\pi/8,-1.178,0.393)$. The corresponding biased gait for Gait 4 is coined by setting the scaling coefficient, $\eta$ in Formula (14) and (15), as 80%.

Figure 21 plots the unacceptable attitudes for the gait on the four-dimensional gait surface (red curves) and for the biased gait (blue curves).

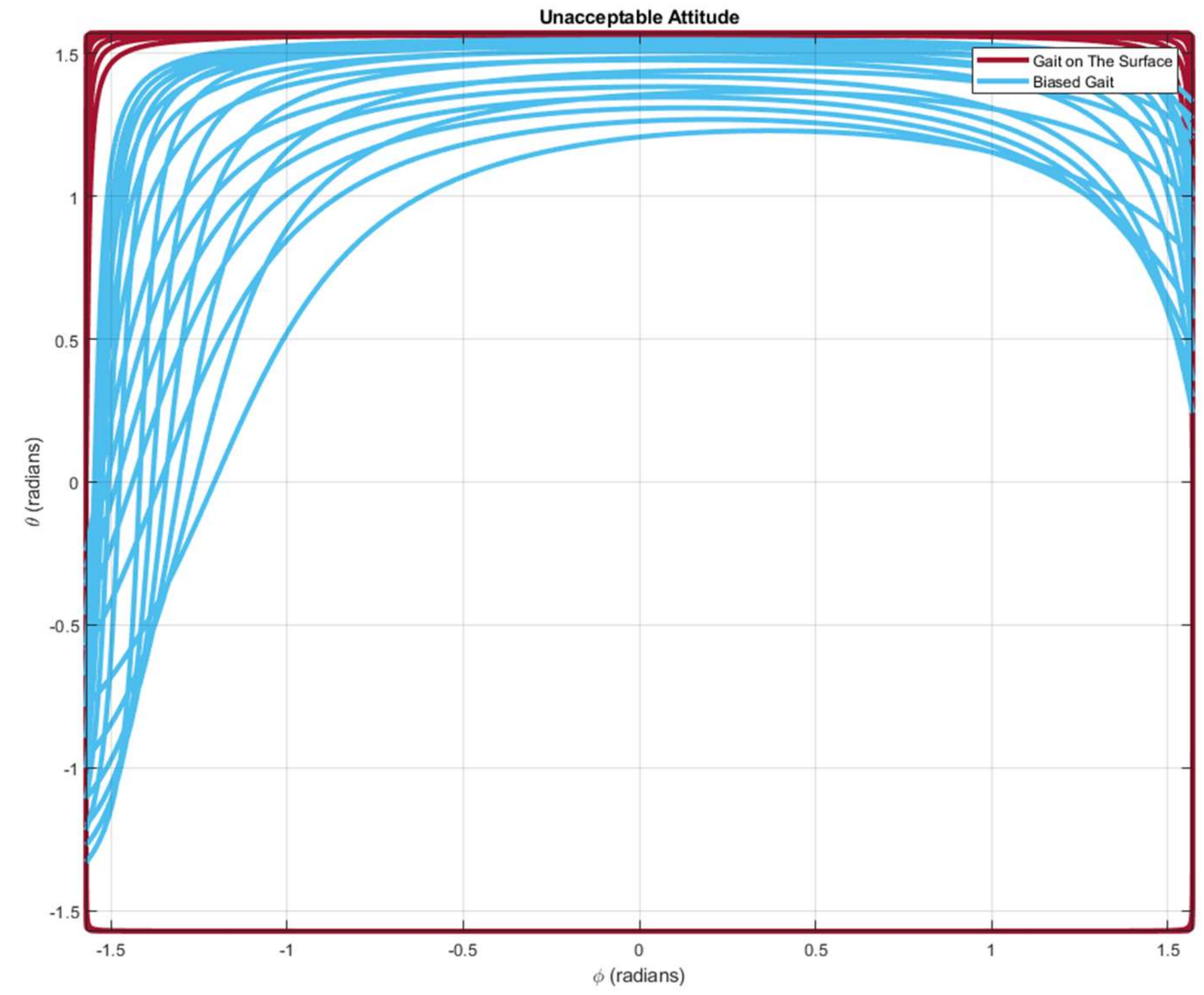

**Figure 21.** The unacceptable attitudes for Gait 4 and for biased Gait 4. The red curve represents the attitudes receiving the singular decoupling matrix in Gait 4. The blue curve represents the attitudes receiving the singular decoupling matrix in biased Gait 4.

Clearly, the red curves are generally further to the origin, $(\phi,\theta) = (0,0)$, comparing with the blue curves, demonstrating that the gait on the four-dimensional gait surface is allowed to move in a wilder region of attitude. The region of admissible attitudes enlarges nearly to the whole region of $\phi \in (-\pi/2,\pi/2), \theta \in (-\pi/2,\pi/2)$.

## 6. Conclusions and Discussions

The proposed four-dimensional gait surface helps plan the gait robust to the attitude change. The gaits on the four-dimensional gait surface show the wider region of acceptable attitudes comparing with the relevant gait biased by partially-scaling.

Although scaling has been proved to be a valid approach in modifying an unacceptable gait which results in a singular decoupling matrix, partially-scaling may not increase the robustness and even have the opposite effect for some gaits.

Two Color Map Theorem assists finding the continuous gait on the four-dimensional gait surface. Multiple gaits can be found on the four-dimensional gait surface without violating this theorem.

The robust gaits distributed on the four-dimensional gait surface can be classified into $(+,+)$ type and $(-,-)$ type.

The gait planned on the planes $(+,+)$ generally receives a larger acceptable attitude zone than the gait planned on the planes $(-,-)$.

An unusual result in Gait 2 is worth discussing.

Figure 22 plots the unacceptable attitudes for Gait 2 on the four-dimensional gait surface (red curves) and for the corresponding biased Gait 2 by scaling with $\eta = 80\%$ (blue curves).

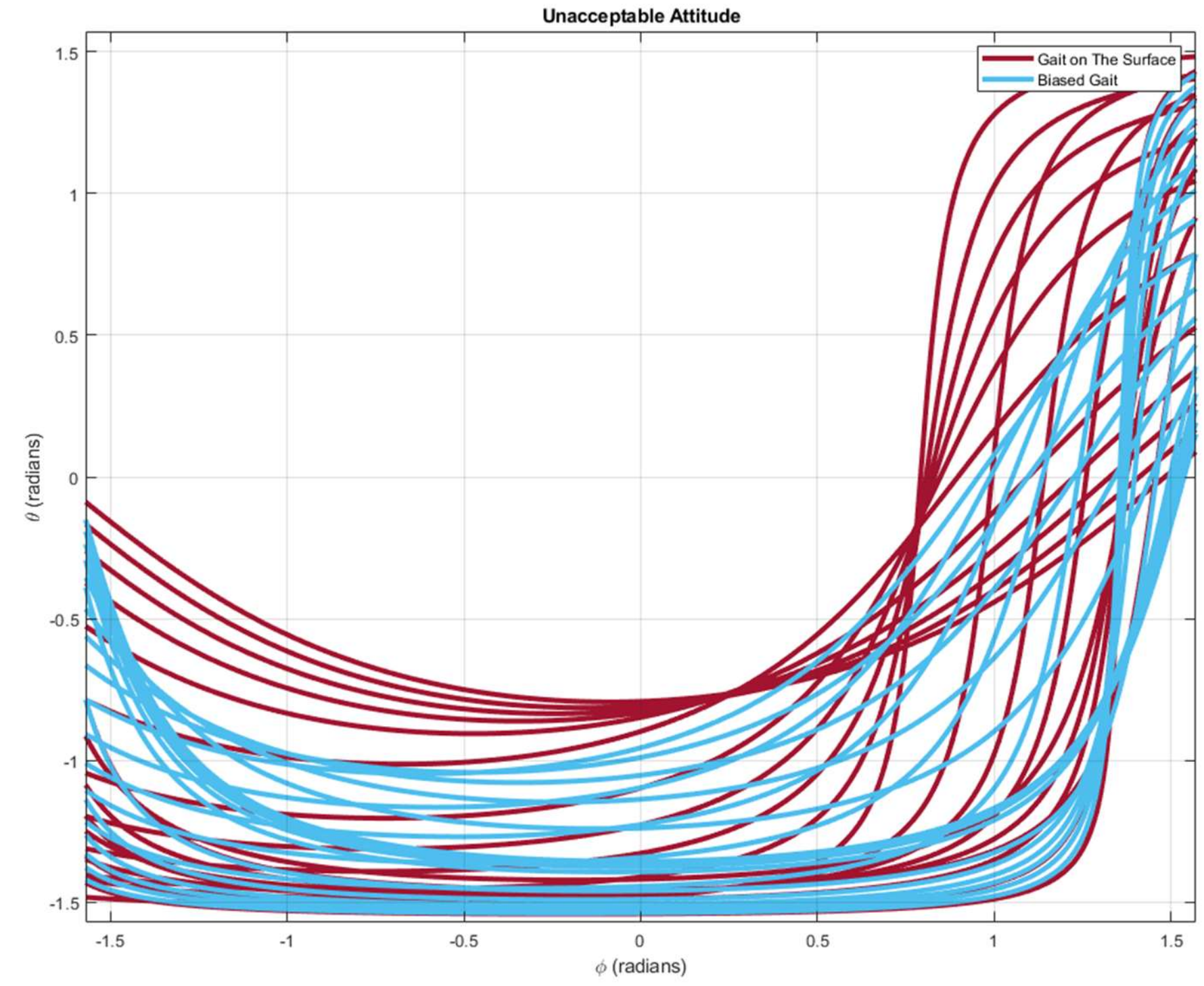

**Figure 22.** The unacceptable attitudes for Gait 2 and for biased Gait 2 ($\eta$ = 80%). The red curve represents the attitudes receiving the singular decoupling matrix in Gait 2. The blue curve represents the attitudes receiving the singular decoupling matrix in biased Gait 2 ($\eta$ = 80%).

It can be seen that the robustness of Gait 2 has little advantages comparing with this biased gait by scaling ($\eta$ = 80%). While it shows advantage comparing with the other biased gait ($\eta$ = 99%) in Figure 19. This may be caused by the fact that Gait 2 is a locally most robust gait.

There are several further research. Firstly, applying the gait into the feedback linearization in a tilt-rotor control is desired, e.g., a tracking experiment with different gaits. Secondly, the underlying mechanism that the gaits on the planes (+,+) receives greater robustness comparing with the gaits on the planes (−,−) is waiting to be elucidated.

## References

1. Chen, C.-C.; Chen, Y.-T. Feedback Linearized Optimal Control Design for Quadrotor With Multi-Performances. *IEEE Access* **2021**, *9*, 26674–26695, doi:10.1109/ACCESS.2021.3057378.
2. Lian, S.; Meng, W.; Lin, Z.; Shao, K.; Zheng, J.; Li, H.; Lu, R. Adaptive Attitude Control of a Quadrotor Using Fast Nonsingular Terminal Sliding Mode. *IEEE Transactions on Industrial Electronics* **2022**, *69*, 1597–1607, doi:10.1109/TIE.2021.3057015.
3. Chang, D.E.; Eun, Y. Global Chartwise Feedback Linearization of the Quadcopter With a Thrust Positivity Preserving Dynamic Extension. *IEEE Trans. Automat. Contr.* **2017**, *62*, 4747–4752, doi:10.1109/TAC.2017.2683265.
4. Martins, L.; Cardeira, C.; Oliveira, P. Feedback Linearization with Zero Dynamics Stabilization for Quadrotor Control. *J Intell Robot Syst* **2021**, *101*, 7, doi:10.1007/s10846-020-01265-2.
5. Kuantama, E.; Tarca, I.; Tarca, R. Feedback Linearization LQR Control for Quadcopter Position Tracking. In Proceedings of the 2018 5th International Conference on Control, Decision and Information Technologies (CoDIT); IEEE: Thessaloniki, April 2018; pp. 204–209.

6. Ansari, U.; Bajodah, A.H.; Hamayun, M.T. Quadrotor Control Via Robust Generalized Dynamic Inversion and Adaptive Non-Singular Terminal Sliding Mode. *Asian Journal of Control* **2019**, *21*, 1237–1249, doi:10.1002/asjc.1800.
7. Lewis, F.; Das, A.; Subbarao, K. Dynamic Inversion with Zero-Dynamics Stabilisation for Quadrotor Control. *IET Control Theory & Applications* **2009**, *3*, 303–314, doi:10.1049/iet-cta:20080002.
8. Taniguchi, T.; Sugeno, M. Trajectory Tracking Controls for Non-Holonomic Systems Using Dynamic Feedback Linearization Based on Piecewise Multi-Linear Models. **2017**, 13.
9. Lee, D.; Jin Kim, H.; Sastry, S. Feedback Linearization vs. Adaptive Sliding Mode Control for a Quadrotor Helicopter. *Int. J. Control Autom. Syst.* **2009**, *7*, 419–428, doi:10.1007/s12555-009-0311-8.
10. Mutoh, Y.; Kuribara, S. Control of Quadrotor Unmanned Aerial Vehicles Using Exact Linearization Technique with the Static State Feedback. *JOACE* **2016**, 340–346, doi:10.18178/joace.4.5.340-346.
11. Shen, Z.; Tsuchiya, T. Singular Zone in Quadrotor Yaw–Position Feedback Linearization. *Drones* **2022**, *6*, 20, doi:doi.org/10.3390/drones6040084.
12. Voos, H. Nonlinear Control of a Quadrotor Micro-UAV Using Feedback-Linearization. In Proceedings of the 2009 IEEE International Conference on Mechatronics; IEEE: Malaga, Spain, 2009; pp. 1–6.
13. Ryll, M.; Bulthoff, H.H.; Giordano, P.R. Modeling and Control of a Quadrotor UAV with Tilting Propellers. In Proceedings of the 2012 IEEE International Conference on Robotics and Automation; IEEE: St Paul, MN, USA, May 2012; pp. 4606–4613.
14. Rajappa, S.; Ryll, M.; Bulthoff, H.H.; Franchi, A. Modeling, Control and Design Optimization for a Fully-Actuated Hexarotor Aerial Vehicle with Tilted Propellers. In Proceedings of the 2015 IEEE International Conference on Robotics and Automation (ICRA); IEEE: Seattle, WA, USA, May 2015; pp. 4006–4013.
15. Systems Engineering Department, KFUPM, Dhahran 31261, Saudi Arabia; Saif, A.-W.A. Feedback Linearization Control of Quadrotor with Tiltable Rotors under Wind Gusts. *Int. j. adv. appl. sci.* **2017**, *4*, 150–159, doi:10.21833/ijaas.2017.010.021.
16. Mistler, V.; Benallegue, A.; M'Sirdi, N.K. Exact Linearization and Noninteracting Control of a 4 Rotors Helicopter via Dynamic Feedback. In Proceedings of the Proceedings 10th IEEE International Workshop on Robot and Human Interactive Communication. ROMAN 2001 (Cat. No.01TH8591); IEEE: Paris, France, 2001; pp. 586–593.
17. Ryll, M.; Bulthoff, H.H.; Giordano, P.R. A Novel Overactuated Quadrotor Unmanned Aerial Vehicle: Modeling, Control, and Experimental Validation. *IEEE Trans. Contr. Syst. Technol.* **2015**, *23*, 540–556, doi:10.1109/TCST.2014.2330999.
18. Scholz, G.; Trommer, G.F. Model Based Control of a Quadrotor with Tiltable Rotors. *Gyroscopy Navig.* **2016**, *7*, 72–81, doi:10.1134/S2075108716010120.
19. Imamura, A.; Miwa, M.; Hino, J. Flight Characteristics of Quad Rotor Helicopter with Thrust Vectoring Equipment. *Journal of Robotics and Mechatronics* **2016**, *28*, 334–342, doi:10.20965/jrm.2016.p0334.
20. Ryll, M.; Bulthoff, H.H.; Giordano, P.R. First Flight Tests for a Quadrotor UAV with Tilting Propellers. In Proceedings of the 2013 IEEE International Conference on Robotics and Automation; IEEE: Karlsruhe, Germany, May 2013; pp. 295–302.
21. Kumar, R.; Nemati, A.; Kumar, M.; Sharma, R.; Cohen, K.; Cazaurang, F. Tilting-Rotor Quadcopter for Aggressive Flight Maneuvers Using Differential Flatness Based Flight Controller.; American Society of Mechanical Engineers: Tysons, Virginia, USA, October 11 2017; p. V003T39A006.
22. Shen, Z.; Tsuchiya, T. Gait Analysis for a Tiltrotor: The Dynamic Invertible Gait. *Robotics* **2022**, *11*, 33, doi:10.3390/robotics11020033.
23. Shen, Z.; Ma, Y.; Tsuchiya, T. Feedback Linearization Based Tracking Control of A Tilt-Rotor with Cat-Trot Gait Plan. *arXiv:2202.02926 [cs.RO]* **2022**.
24. Altug, E.; Ostrowski, J.P.; Mahony, R. Control of a Quadrotor Helicopter Using Visual Feedback. In Proceedings of the Proceedings 2002 IEEE International Conference on Robotics and Automation (Cat. No.02CH37292); May 2002; Vol. 1, pp. 72–77 vol.1.

25. Zhou, Q.-L.; Zhang, Y.; Rabbath, C.-A.; Theilliol, D. Design of Feedback Linearization Control and Reconfigurable Control Allocation with Application to a Quadrotor UAV. In Proceedings of the 2010 Conference on Control and Fault-Tolerant Systems (SysTol); IEEE: Nice, France, October 2010; pp. 371–376.
26. Mokhtari, A.; Benallegue, A. Dynamic Feedback Controller of Euler Angles and Wind Parameters Estimation for a Quadrotor Unmanned Aerial Vehicle. In Proceedings of the IEEE International Conference on Robotics and Automation, 2004. Proceedings. ICRA '04. 2004; IEEE: New Orleans, LA, USA, 2004; pp. 2359-2366 Vol.3.
27. Shen, Z.; Tsuchiya, T. Cat-Inspired Gaits for A Tilt-Rotor -- from Symmetrical to Asymmetrical. *arXiv:2203.12057 [cs.RO]* **2022**, 12.
28. Wisleder, D.; Zernicke, R.F.; Smith, J.L. Speed-Related Changes in Hindlimb Intersegmental Dynamics during the Swing Phase of Cat Locomotion. *Exp Brain Res* **1990**, *79*, doi:10.1007/BF00229333.
29. Verdugo, M.R.; Rahal, S.C.; Agostinho, F.S.; Govoni, V.M.; Mamprim, M.J.; Monteiro, F.O. Kinetic and Temporospatial Parameters in Male and Female Cats Walking over a Pressure Sensing Walkway. *BMC Veterinary Research* **2013**, *9*, 129, doi:10.1186/1746-6148-9-129.
30. Appel, K.; Haken, W. The Solution of the Four-Color-Map Problem. *Scientific American* **1977**, *237*, 108–121.
31. Franklin, P. The Four Color Problem. *American Journal of Mathematics* **1922**, *44*, 225–236, doi:10.2307/2370527.